\documentclass[runningheads]{llncs}
\usepackage{graphicx}
\usepackage{amsmath,amssymb} 
\usepackage{color}
\usepackage{subfig}
\usepackage{booktabs}
\usepackage{multirow}
\usepackage[width=122mm,left=12mm,paperwidth=146mm,height=193mm,top=12mm,paperheight=217mm]{geometry}
\begin{document}
\pagestyle{headings}
\mainmatter
\def\ECCV18SubNumber{***}  

\title{Self-Supervised Monocular Image Depth Learning and Confidence Estimation} 

\titlerunning{Self-Supervised Depth Learning \& Confidence Estimation}

\authorrunning{Chen \textit{et al}}

\author{Long Chen\inst{1}, Wen Tang\inst{1}, Nigel John\inst{2}}
\institute{Bournemouth University, Poole, UK\\
\and
University of Chester, Chester, UK\\
\email{chenl@bournemouth.ac.uk}}

\maketitle

\begin{abstract}
Convolutional Neural Networks (CNNs) need large amounts of data with ground truth annotation, which is a challenging problem that has limited the development and fast deployment of CNNs for many computer vision tasks. We propose a novel framework for depth estimation from monocular images with corresponding confidence in a self-supervised manner. A fully differential patch-based cost function is proposed by using the Zero-Mean Normalized Cross Correlation (ZNCC) that takes multi-scale patches as a matching strategy. This approach greatly increases the accuracy and robustness of the depth learning. In addition, the proposed patch-based cost function can provide a 0 to 1 confidence, which is then used to supervise the training of a parallel network for confidence map learning and estimation. Evaluation on KITTI dataset shows that our method outperforms the state-of-the-art results.

\end{abstract}

\section{Introduction}

The human vision system is amazingly complex and extremely delicate. It can perceive depth through stereopsis, which relies on the displacement of the same object between the images received by the left and right retinas \cite{Dunkin2015}. With extensive visual experience and through trial and error, humans develop the ability to use contextual depth cues to achieve good and reliable perception of depth and better understanding of spatial structure. Among these depth cues, some of them do not rely on stereopsis, such as object occlusion, perspective, familiar and relative size, depth from motion, lighting and shading. 
Therefore, if blind in one eye or if performing a monocular task such as endoscopic surgery, we can still judge distance from these many different intuitive depth cues. In contrast, when using machine vision it is hard to infer the non-stereopsis depth cues. With the recent development of Deep Convolutional Neural Networks (DCNNs), machines can solve many computer vision problems when provided with very large human annotated datasets such as ImageNet \cite{Krizhevsky2012}, which is known as supervised learning. Acquisition of labelled datasets is one of the biggest challenges for supervised learning, however, which is an expensive, time-consuming and labour-intensive task. 
\begin{figure}
\centering
\includegraphics[width=0.98\textwidth]{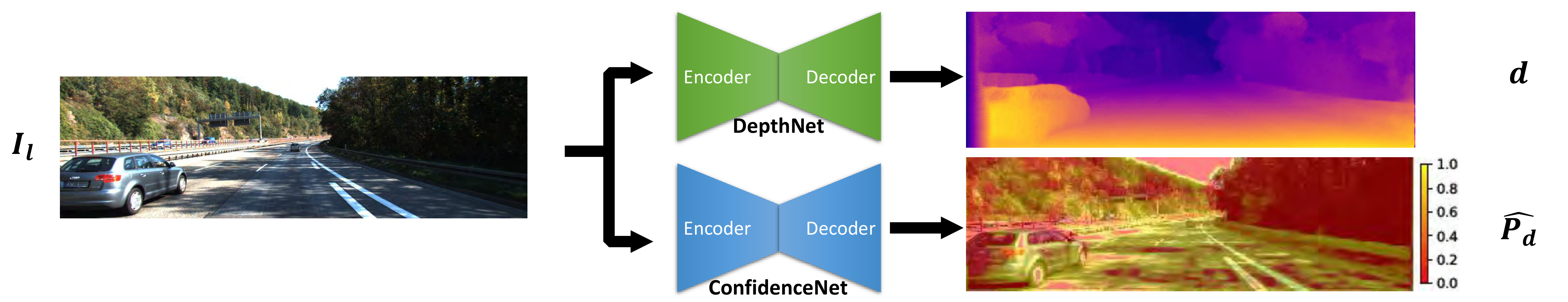}
\caption{Our proposed framework can simultaneously estimate depth and the confidence of estimated depth. }
\label{intro}       
\end{figure}
In this paper, we propose a novel self-supervised computational framework that mimics the process of how a human learns varies of contextual depth cues from stereopsis. We train a DCNN for synthesizing depth from one view of the stereo image pair, then reconstruct the other view by the synthesized depth, and finally using the stereo vision epipolar constraint \cite{Zhang1998} to minimize the error of the depth synthesis. 

Our approach does not require the ground truth depth for supervised training. Instead, we derive the implicit function of estimating depth from monocular images by the epipolar constraint of the stereo image pair. Therefore, the method can be regarded as self-supervised learning. Compared with previous work \cite{Garg2016} \cite{Godard2017} \cite{Zhou2017} addressing the same problem, we incorporate a patch-based image evaluation strategy, inspired by the classic patch matching algorithms for finding the best-matched patches between the left and right images. We use the Zero-Mean Normalized Cross Correlation (ZNCC) to measure the normalized similarities between these patches. A fully-differential patch-based ZNCC cost function is implemented to guide the depth synthesis process for more accurate results. Visual assessment shows that our approach can produce more accurate and robust depth estimations in both texture-rich and texture-less areas due to the enlargement of matching field from a pixel to a patch (see Figure \ref{l1_zncc}). Empirical evaluations on KITTI dataset demonstrate the effectiveness of our approach and produce a state-of-the-art performance in monocular depth estimation task.

Our second contribution is that we train a parallel DCNN to evaluate the performance of the monocular depth estimation and output a 0 to 1 confidence map. The parallel DCNN is also trained in a self-supervised manner thanks to our ZNCC similarity measurement function. As ZNCC is a normalized measure of similarity, which can be approximated as the confidence of the depth estimation, we take the ZNCC loss to self-supervise the parallel DCNN (ConfidenceNet) during training so that we can estimate the confidence of the depth estimated from the first DCNN (DepthNet) during testing mode as shown in Figure \ref{intro}. A confidence map is extremely useful for the monocular depth estimation task trained in an unsupervised manner, as the learned epipolar constraint only works well when there are clear corresponding pixels between the image pairs; it will fail and produce uncertain depth when occlusion and specularity exist in images. Our confidence map can give a basic assessment of the reliability of the predicted depth, which can then be further integrated into many applications such as monocular dense reconstruction, SLAM-based depth fusion \cite{Tateno2017}, and many tasks need crucial accurate and confidence such as monocular endoscopic surgery.

\section{Related Work}

\subsubsection{Stereo Depth Estimation.}
The problem of stereo images depth estimation has been well studied for a long time \cite{Barnard1982} \cite{Scharstein2001}. With the theory of epipolar constraint, accessing depth from stereo images can be regarded as a well-posed problem when ignoring the occlusions and depth discontinuities. Many stereo vision algorithms managed to achieve comparable results to ground truth depth acquired from depth sensors \cite{Hirschmuller2008} \cite{Kendall2017}.

\subsubsection{Monocular Depth Estimation.}
In contrast, estimating depth from monocular images is an ill-posed problem that is inherently ambiguous \cite{Eigen2014}, and many research efforts have been devoted to the problem of monocular image depth estimation. One of the classic methods is Shape from Shading (SFS) \cite{Zhang1999}, which is based on the gradual variation of shading as a cue to estimate the shape and depth. However, SFS has a strict prior assumption of Lambertian reflectance, uniform color and texture, and fixed light source direction, which are not applicable to most of the images in the real world. Saxena et al \cite{Saxena2006}\cite{Saxena2007}\cite{Saxena2008}\cite{Saxena2009} used Markov Random Field (MRF) incorporated with multiscale image features to learn monocular cues in a supervised manner. However, the hand-craft local features used in these approaches limit the expressive power of supervised learning, and lack a global contextual understanding of the scene for learning consistent depth. 

\subsubsection{DCNNs based Monocular Depth Learning.}
More recently, DCNNs \cite{Eigen2014} \cite{Eigen2015} are introduced to solve the challenge of monocular depth estimation problem, and has pushed the state-of-the-art forward in this area. Building on the success of this approach, several improvements have been made by incorporating probabilistic models such as Conditional Random Fields (CRFs)\cite{Li2015} \cite{Liu2014} \cite{Liu2016} \cite{Xu2017}, advanced network structures such as Resnet \cite{Laina2016}, two-streamed networks \cite{Li2017}, multi-task joint training \cite{Ladicky2014} \cite{Eigen2015} \cite{Wang2015} \cite{Mousavian2016} and novel loss functions such as sparse supervision \cite{Kuznietsov2017}, relative depth \cite{Zoran2015}\cite{Chen2016} and depth as classification \cite{Cao2017}. Impressive as these works are, ground-truth depth data are still needed for the supervision of training these DCNNs.

\subsubsection{Unsupervised Monocular Depth Learning.}
Driven by DCNNs, view synthesis technology \cite{Fitzgibbon2003} has proven to be effective on synthesizing new views by sampling pixels from existing views \cite{Zhou2016} \cite{Flynn2016}, which enables novel frameworks of unsupervised learning of monocular depth from stereo pairs, e.g., Deep3D \cite{Xie2016}, Garg \textit{et al} \cite{Garg2016}. The works by Godard \textit{et al} \cite{Godard2017} and Zhou \textit{et al} \cite{Zhou2017} advanced the networks by incorporating left-right consistency and pose estimations. However, a common weakness of these approaches is the use of pixel-wised photometric loss (L1-norm) to construct loss functions to guide the back-propagation process. Gradients are derived from the pixel intensity difference \cite{Zhou2017}, which will lead to ambiguous gradients in texture-less areas and also in the regions that contain the mixture of thin structures and texture-less areas. Although multi-scale and smoothness loss functions are used to prevent such issue \cite{Garg2016} \cite{Godard2017} \cite{Zhou2017}, the result is still not desirable and gradients are still likely to converge to local minimums due to the ambiguous pixel-wise loss. As shown in Figure \ref{l1_zncc}, in a common speed limitation board area from the KITTI dataset, the direct pixel-wise photometric loss will lead to many local minimums shown in the right curve chart. While as the left curve chart shows the result of using our proposed patch-based ZNCC loss, the loss is more smooth and likely to converge to the global minimum. And the experiment result (the last row in Figure \ref{l1_zncc}) shows our proposed method can effectively generate accurate depth in complex regions.

\subsubsection{Novelty Compared to Previous Work.}
 We propose a novel multi-scale patch-based cost function that adopts the ZNCC as a similarity function to explicitly enlarge the matching field and increase the matching robustness. From another point of view, our proposed patch-based cost function implicitly integrate the classic Patch Matching (PM) algorithm as a minimization problem in our loss function. Although Garg \textit{et al} \cite{Garg2016} have discussed a straightforward idea of using the stereo matching algorithm as a pre-processing method to generate ''quasi ground-truth'' depth for training, their result is not desirable due to the poor quality of ''quasi ground-truth''. Recently, Luo \textit{et al} \cite{Luo2018} also proposed a similar framework that firstly use a DCNN to synthesize stereo pairs from single images, and then use stereo matching to get depth. In contrast to these two works, we treat the stereo matching as a minimization problem and implement a fully differential PM algorithm as a cost function that is seamlessly integrated into our neural network. As the loss of the PM cost function can be passed through the whole network during a backward propagation, our network can produce more robust and consistent depth by large-scale self-supervised training, which will not be limited by the performance of off-the-shelf stereo matching algorithms.
 
 Another novelty of our work is the confidence map. As monocular depth estimation itself is an ill-posed problem, although learning-based approaches achieve comparable results to stereo depth estimation, there are still many unavoidable mistakes in the predicted depth map. For the first time, our method is able to provide a pixel-wise confidence of the predicted depth by using a parallel DCNN to capture and learn the confidence during training. The confidence map will greatly improve the usability of deploying monocular depth estimation into many practical tasks. 

\section{Method}
\label{Method}

\subsection{Framework Overview}

\begin{figure}
\centering
\includegraphics[width=0.98\textwidth]{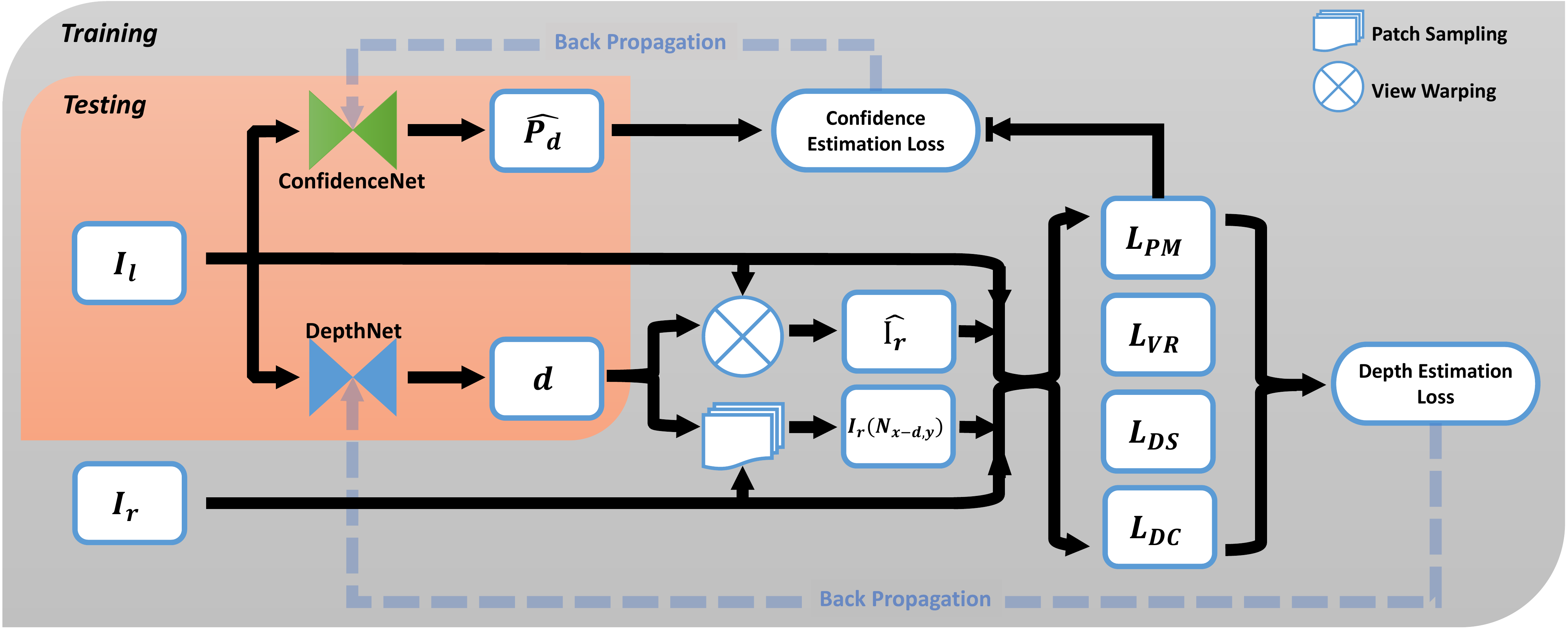}
\caption{Framework for proposed self-supervised monocular depth learning and confidence estimating networks. }
\label{overview}       
\end{figure}

Figure \ref{overview} illustrates the entire framework for our self-supervised monocular depth learning and confidence estimation networks. Since the ground-truth depth $D_{gt}$ is absent for supervised training, we treat the monocular depth estimation as a problem of image synthesis error minimization during training. Specifically, during training, we use the left images $I_{l}$ of the stereo pairs to synthesize per-pixel depth $D$ using an encoder-decoder network $D=F_{depth}(I_{l},\theta)$, which is converted into disparities maps $d$ by the Equation \ref{triangulation}. The disparities map $d$ is then used to guide the stereo view reconstruction $\hat{I_{r}}=F_{warp}(I_{l}, d)$ and the sampling of patches $N_{x-d,y}=F_{sample}(I_{r}, d)$. After that, the loss function  $L_{total}$ is calculated based on Patch Matching Loss $L_{PM}$, View Reconstruction Loss $L_{VR}$, Disparity Smoothness Loss $L_{DS}$, and Disparity Consistency Loss $L_{DC}$. As these processes are differentiable, back propagation can be used to update the parameters $\theta$ of our depth learning network to minimize the total loss $L_{total}$.

\begin{align}
    \frac{\partial L_{total}}{\partial \theta} = &\frac{\partial L_{PM}+\partial L_{VR}+\partial L_{DS}+\partial L_{DC}}{\partial F_{warp}(I_{l}, d)+\partial F_{sample}(I_{r}, d)} \times \frac{\partial F_{warp}(I_{l}, d)+\partial F_{sample}(I_{r}, d)}{\partial d} \\ 
    &\times \frac{\partial d}{\partial D}\times \frac{\partial D}{\partial F_{depth}(I_{l},\theta)}\times \frac{\partial F_{depth}(I_{l},\theta)}{\partial \theta}\nonumber
\end{align}

Since our patch-based ZNCC loss map $L_{PM}(x,y)$ represents the normalized inverted similarity between each pixel of the $I_{l}$ and $I_{r}$, it can be approximated as the inverted confidence of the depth estimation result. We use the $L_{PM}(x,y)$ to self-supervise the training of a second encoder-decoder network -- ConfidenceNet to generate the confidence $\hat{P_{d}}$ of the per-pixel depth estimation of our DepthNet.

\subsection{Depth Synthesis Network}

\begin{figure}
\centering
\includegraphics[width=0.9\textwidth]{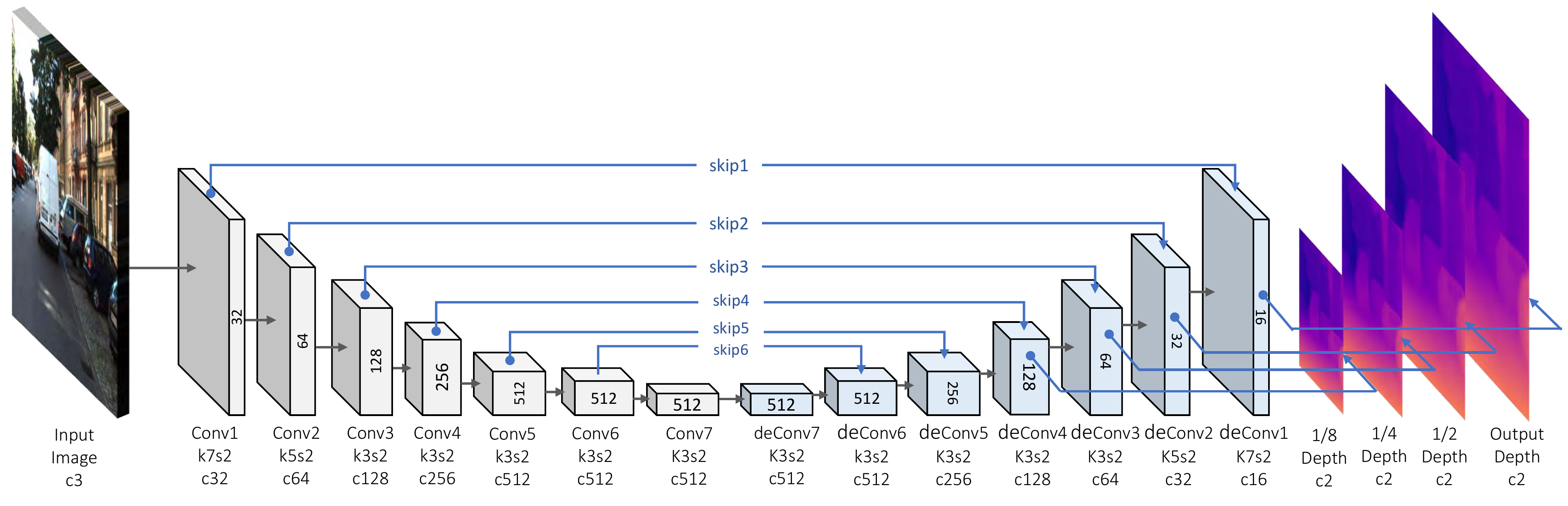}
\caption{Depth synthesis network structure. ''k'' is the kernel size, ''s'' for the stride, ''c'' for the channel number. For simplicity, we do not draw the conv layers after each conv and deconv layer, which have the same kernel and channel size as previous layers but with stride 1. }
\label{network1}       
\end{figure}

The core part of our framework is the depth synthesis and generation. Our goal is to learn an implicit function $F_{depth}$ that estimates a per-pixel depth from a single input image. Inspired by the architectures of FlowNet \cite{Dosovitskiy2017}, DispNet \cite{Mayer2016} and the network of Godard \textit{et al} \cite{Godard2017} and Zhou \textit{et al} \cite{Zhou2017}, we employ a VGG-like fully convolutional neural network architecture \cite{Shelhamer2017} in order to generate per-pixel depth from a single image. Our encoder-decoder model is illustrated in Figure \ref{network1}. The input image is encoded by 7 conv layers with stride 2 each followed by a conv layer with stride 1, which efficiently compress the input image into a feature tensor with $1/2^{7}$ original size and 512 channels. Then, the feature tensor is up-sampled by 7 deConv layers with stride 2 each followed by a conv layer with stride 1, which decode the feature tensor into a full original size depth. Following the method in \cite{Dosovitskiy2017}, 6 skip connections are implemented for preserving high-level information to ensure the high quality per-pixel prediction after up-sampling. Multi-scale depth images are outputted and used for further steps to constraint the network for a coarse-to-fine up-sampling.

\subsection{Warping-based Stereo View Reconstruction}
\begin{figure}
\centering
\subfloat[Foward mapping]{\includegraphics[width=0.48\textwidth]{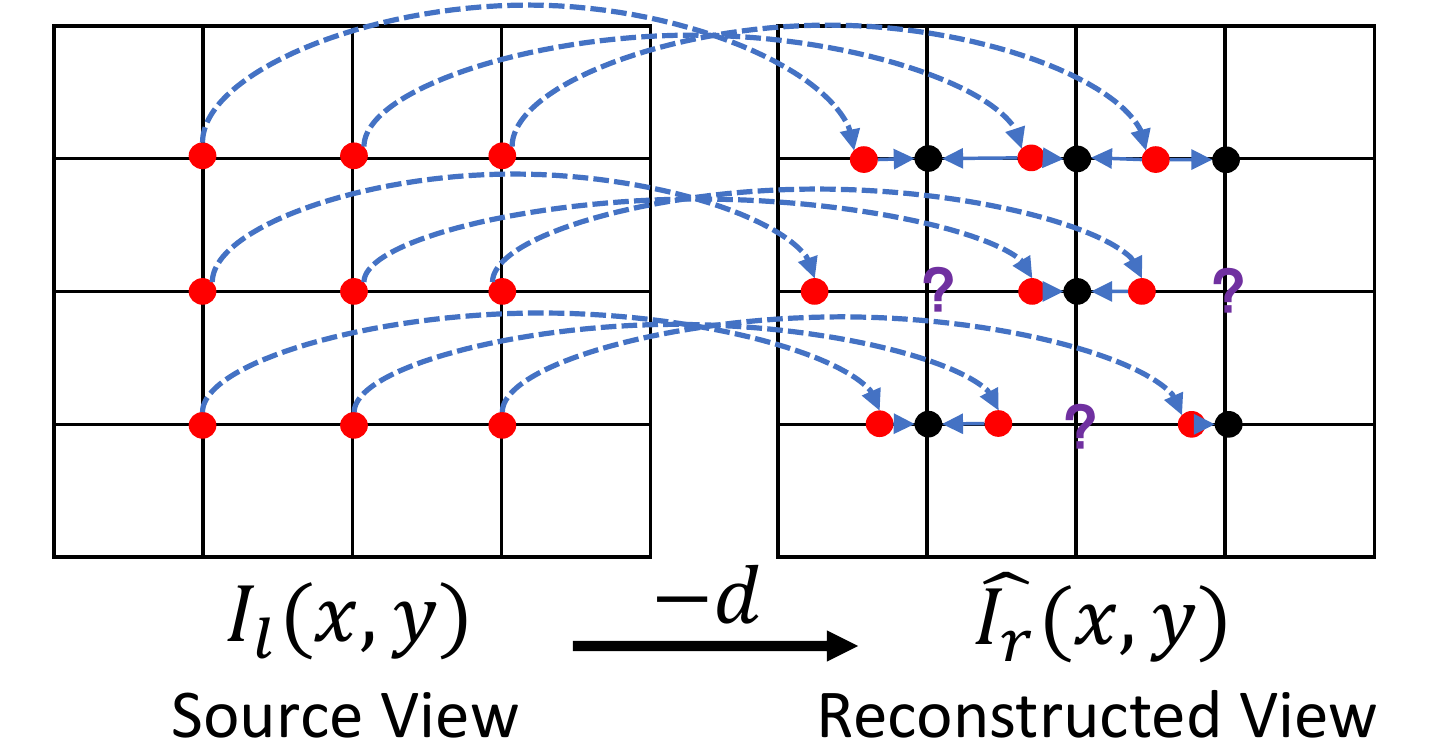}}\
\subfloat[Backward mapping]{\includegraphics[width=0.48\textwidth]{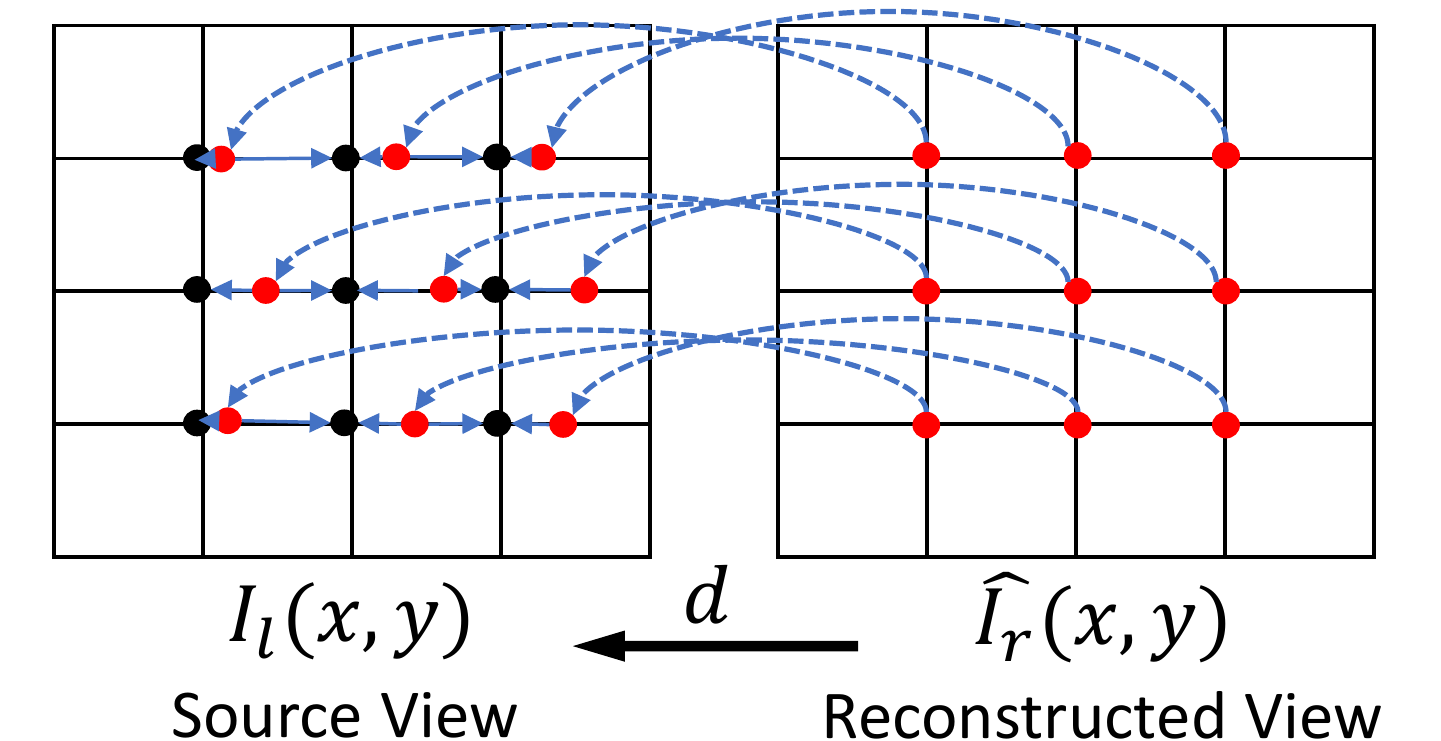}}\\
\caption{The difference between forward mapping and backward mapping.}
\label{fmbm}       
\end{figure}

View warping is an enabling technology for self-supervised learning framework \cite{Garg2016} \cite{Godard2017} \cite{Zhou2017}. Given the per-pixel disparity map estimated from a single image in the previous step, the target view of the stereo pairs can be reconstructed by the epipolar relationship in stereo vision. According to the epipolar constraint: the projection of a pixel $x_{l}$ on the right camera plane $x_{r}$ must be contained in the epipolar line. For calibrated stereo pairs discussed in this paper, $x_{l}$ and $x_{r}$ must be in the same row $y$, and the disparity $d$ describes the horizontal displacement of the corresponding pixels $x_{l}$ and $x_{r}$ . Through the stereo triangulation, we can get that
\begin{equation}
\label{triangulation}
    D_{xy} = \frac{bf}{d} \; \Rightarrow \;  d= x_{l}-x_{r} = \frac{bf}{D_{xy}}
\end{equation}
where $D_{xy}$ is the depth estimated in the pixel at $(x,y)$, b and f are the camera baseline and focal distance.
By the relationship discussed in the above equation, the target view in a stereo pair can be reconstructed given the source view and the corresponding depth (estimated through our depth synthesis network). 

However, the direct mapping from one known view to the other view (forward mapping) will result in holes in the target image that are not differentiable. Therefore, we use the inverse mapping: for each pixel in the target view, by picking points from the source to reconstruct the target view guided by the $d$. Thus, a complete and differentiable target view can be generated. Then the bilinear sampling \cite{Jaderberg2015} is used to get the interpolated pixel value from the source view.

\subsection{Disparity-guided Patch Sampling}
Inspired by the stereo view reconstruction described above, we propose a novel patch sampling process guided by the estimated disparity from our DepthNet. $N_{x,y}$ is defined as a patch with window size $n$, centered at the coordinate $(x,y)$. We sample patches on each pixel in the left image $\left \{ x,y\in I_{l}| I_{l}(N_{x,y}) \right \}$, and the corresponding patches shifted by disparity values $d$ of each pixel in the right image, $\left \{ x,y\in I_{r}| I_{r}(N_{x-d,y}) \right \}$. According to Equation \ref{triangulation}, if $d$ is correct, then we have $I_{l}(N_{x,y})=I_{r}(N_{x-d,y})$. And this relationship will be used to construct the patch matching loss. These sampled patches are computed and stored vectorized so that can be deployed parallelly on GPU for accelerated computation. 

The patch sampling size is very important and can affect the final performance of similarity measurement. However, there is no optimal patch size and the performance varies greatly across different images and local details. When small patch size is used, little information will be captured, and the similarity comparison robustness will be decreased. If we use a large patch size, computational complexity will be greatly increased and also cannot recover accurate depth at stereo occlusion and depth discontinuous. Therefore, we use a multi-scale patch sampling scheme and sample a combination of 4 different patch sizes in an image to fully exploit the effects of different patch sizes. We will discuss the choice of patch sizes in Section \ref{multi-scale}.

\subsection{Loss Function Construction}
We define a loss function $L_{total}$ with multiple strategies to effectively train our networks for accurate, smooth and realistic depth. 
\begin{equation}
    L_{total}= \omega_{p} L_{PM}+\omega_{v} L_{VR}+\omega_{d} L_{DS}+ \omega_{c} L_{DC}
\end{equation}
where from left to right is: Patch Matching Loss, View Reconstruction Loss, Disparity Smoothness Loss and Disparity Consistency Loss. $\omega$ is the corresponding weights to balance the effects of gradients back propagation. Each loss function will be explained in details below:

\subsubsection{Patch Matching Loss.}

Inspired by patch matching algorithm that by finding the best-matched patches in the left and right image to get correct disparities. We propose a patch matching loss that maximize the similarities (minimize the differences) of patches in left image $I_{l}(N_{x,y})$ and the shifted patches in right image $I_{r}(N_{x-d,y})$ to get correct disparities. Here, the ZNCC measure of similarity is used to compute a normalized similarity between the patches $I_{l}(N_{x,y})$ and $I_{r}(N_{x-d,y})$:

\begin{equation}
\resizebox{.9\hsize}{!}{$C_{ZNCC}\left ( I_{l}(N_{x,y}),I_{r}(N_{x-d,y})\right )=\frac{\sum_{i,j\in N_{x,y}} \left ( I_{l}\left ( i,j \right )-\bar{I}_{l}\left ( N_{x,y} \right ) \right )\cdot \left ( I_{r}\left ( i-d,j \right )-\bar{I}_{r}\left ( N_{x-d,y}\right ) \right )}{\sqrt{\sum_{i,j\in N_{x,y}}\left ( I_{l}\left ( i,j \right )-\bar{I}_{l}\left ( N_{x,y} \right ) \right )^{2}\cdot \sum_{i,j\in N_{x,y}}\left ( I_{r}\left ( i-d,j \right )-\bar{I}_{r}\left ( N_{x-d,y} \right ) \right )^{2}}}$}
\end{equation}
where $\bar{I}\left ( N_{x,y} \right ) =\frac{1}{n}\sum_{x,y\in N_{x,y}}I\left ( x,y \right )$ is the mean intensity of the patch $N_{x,y}$ centered at the coordinate $(x,y)$. 

The ZNCC returns a similarity ranging from $[-1,1]$. We first normalize it into $[0,1]$ then invert it to get the patch matching loss:

\begin{equation}
\label{lpm}
L_{PM} = \sum_{x,y} 1- \frac{ 1 + C_{ZNCC}\left(I_{l}(N_{x,y}),I_{r}(N_{x-d,y})\right )}{2}
\end{equation}

\begin{figure}[bt]
\centering
\includegraphics[width=0.7\textwidth]{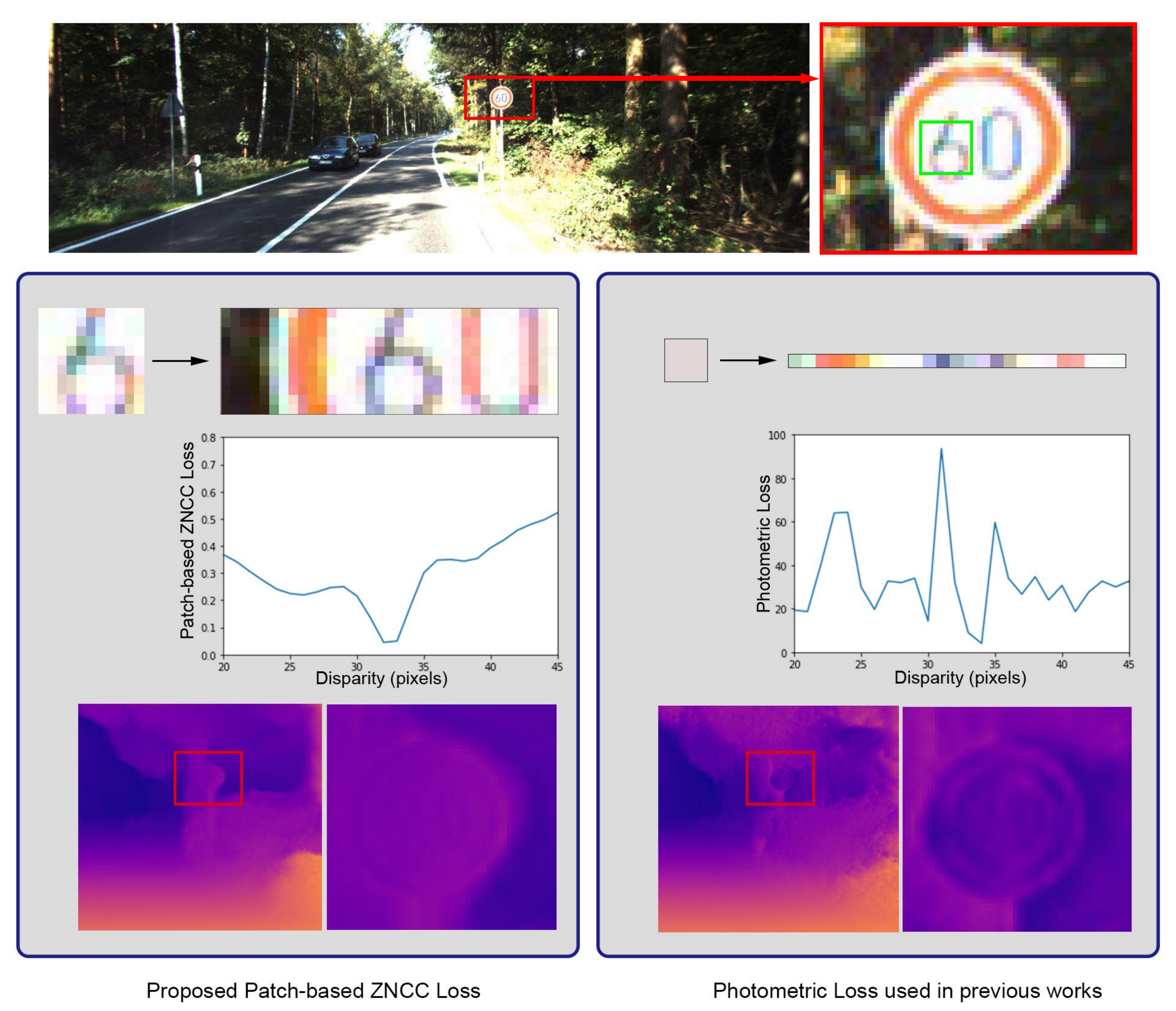}
\caption{Comparison of our proposed patch-based ZNCC loss with the photometric loss used in previous works.}
\label{l1_zncc}       
\end{figure}

Our patch matching loss is computed at all 4 patch sizes to cover both small structures and large areas. There are several advantages of using our patch-based ZNCC loss to regularize the depth synthesis:

(1) Our patch matching loss uses patches for measurement that involve larger regions than the direct pixel-wise photometric loss used in previous work, which is more robust and can achieve sub-pixel accuracy. Figure \ref{l1_zncc} demonstrates the effect of our patch-based ZNCC loss. We charted the values of our patch-based ZNCC loss and the photometric loss against the disparity value of a pixel located at the center of the image patch ''6''. It is obvious that by using our proposed patch-based ZNCC loss, the loss is more smooth and likely to converge to the global minimum. Whereas the direct pixel-wise photometric loss will lead to many local minimums shown in the right curve chart. 

(2) Compared to other similarity measures such as absolute intensity difference (AD), Census, and Normalized Cross Correlation (NCC), ZNCC is especially robust against Gaussian noise and variation between the compared patches, which can help to recover more accurate depth in our self-supervised framework.

(3) As a zero-mean normalized similarity measurement function, our patch-based ZNCC loss can provide a similar value ranging from $[-1,1]$. After normalized to $[0,1]$ as shown in Equation \ref{lpm}, it can be regarded as the confidence of the generated depth at each pixel, which can be further used to self-supervise the training of our confidence network.

\subsubsection{View Reconstruction Loss.}
We use the view reconstruction loss as a second supervision on the depth synthesis. Guided by the synthesized depth, the right views can be reconstructed by collecting pixels from left images. The view reconstruction loss is defined as the L1 loss between the reconstructed view $\hat{I_{r}}$ and the original view $I_{r}$:
\begin{equation}
L_{VR} = \sum_{xy}\left | I_{r}(x,y)-\hat{I_{r}}(x,y)\right |
\end{equation}
Compared to the patch matching loss, the view reconstruction L1 loss is more sensitive to small structures and depth discontinuities and can provide more detailed depth information. 

\subsubsection{Disparity Smoothness Loss.}
We use a disparity smoothness term to regularize our network to produce more smooth depth. Similar to \cite{Garg2016} \cite{Godard2017} \cite{Zhou2017}, we use the sum of the L1 norm of the disparity gradients along the $x$ and $y$ directions as a smoothness factor. The edge-aware terms are used to reduce the penalty on edges where depth discontinuities usually happen, which can prevent over-smoothing.
\begin{equation}
L_{DS} = \frac{1}{xy}\sum_{x,y}\left | \frac{\partial d(x,y)}{\partial x}\right |e^{-\left \| \frac{\partial I(x,y)}{\partial x} \right \|}+ \left | \frac{\partial d(x,y)}{\partial y}\right |e^{-\left \| \frac{\partial I(x,y)}{\partial y} \right \|}
\end{equation}

\subsubsection{Disparity Consistency Loss.}
The left-right disparity consistency loss proposed in \cite{Godard2017} has achieved a great improvement for monocular depth generation. Here, we adopt this loss function into our framework. The left and right image disparities are both generated, and the difference of left disparity map and the reconstructed left disparity map from right disparity is computed and minimized. This loss will ensure the left and right disparities coherence.
\begin{equation}
L_{DC} = \frac{1}{xy}\sum_{x,y}\left |  d_{l}(x,y) - d_{r}(x-d_{l}(x,y),y)\right |
\end{equation}

\subsection{Confidence Estimation Network}
One of the advantages of our proposed patch matching loss is that a normalize similarity measurement can be generated for each pixel at the training time. With the well-known epipolar constraint, the per-pixel confidence of the estimated depth can be approximated as the normalized similarity measurement of the left patches and the corresponding patches in the right image. 

\begin{equation}
P_{d}(x,y) \approx C_{Normalized}(I_{l}(N_{x,y}),I_{r}(N_{x-d,y})) = (1-L_{PM}(x,y))
\end{equation}

Here, we propose to use another encoder-decoder network to learn the confidence map generated by our depth estimation network during training, so that the confidence map can be preserved and generated during the testing time. We tried to train the confidence and depth in one network like \cite{Ladicky2014} \cite{Eigen2015} \cite{Wang2015} \cite{Mousavian2016}, but the multi-task training would reduce the depth estimation performance. Therefore, we use a parallel encoder-decoder network to learn the confidence supervised by the per-pixel ZNCC loss of our depth estimation network. The loss of our ConfidencNet is shown below:
\begin{equation}
L_{ConfidenceNet} = \sum_{x,y}\left | (1-L_{PM}(x,y))-\hat{P_{d}}(x,y)\right |
\end{equation}
where $\hat{P_{d}}(x,y)$ is the generated confidence map, $L_{PM}(x,y)$ is the patch matching loss from our depth estimation network described in above sections. The static copy is used here to prevent the gradients propagating back to the depth estimation network. The $1-L_{PM}(x,y)$ operation inverts the loss to confidence, and L1 loss is used to access the confidence estimation error.

Instead of using the same encoder-decoder network structure as our DepthNet, we employ a simpler structure by only using first 5 conv-layer and last 5 deconv-layer without skip layers as described in Figure \ref{network1} for two reasons: 

(1) To reduce memory usage and training time, as training two neural networks at the same time is very computationally expensive. The second network can be replaced by a deeper and more complex encoder-decoder network to produce sharper and more accurate confidence, but the main purpose of our work is to prove that our self-supervised monocular depth learning and confidence estimation framework is feasible and helpful for depth prediction, hence we choose to use a simple network structure as the proof of concept. 

(2) We intend to use a simpler network with fewer weights to prevent over-fitting to noises and to learn more generic confidence -- high confidence in texture-rich areas, low confidence in texture-less, blurry and occluded areas, which is what we design this confidence net for.

\section{Experiments}

In this section, we evaluate our framework and compare the results with prior approaches both quantitatively and qualitatively on KITTI dataset. We use the rectified stereo image pairs for training our networks. For testing time, we use the left image to generate depth, and the corresponding sparse LIDAR data is served as the ground truth for benchmarking.

\subsection{Implementation Details.}
Our networks are implemented in Tensorflow and trained on a workstation with a single Nvidia Titan X GPU (12G Memory). Our models take around 60 hours to train for 50 epochs. When in testing mode, our networks can output depth and confidence map at around 20 frames per second. 

\textbf{Hyper Parameters.} 
All input images are scaled to 512x256 with a batch size of 4. Adam Optimizer is used with $\beta_{1}  =0.9$, $\beta_{1}  =0.999$, and initial learning rate $\lambda=0.0001$ that decays after half of the training process. The weights to construct our total loss function for depth estimation network are $w_{p}=0.5$,$w_{v}=1$,$w_{d}=0.1$,$w_{c}=1$.

\textbf{Data Augmentation.}
The same data augmentation approach in \cite{Godard2017} is used to randomly flip the image and change the gamma, brightness, and color shifts to increase the network robustness and prevent over-fitting.

\textbf{Multi-scale Implementation.}
\label{multi-scale}
We employ a multi-scale strategy to ensure a coarse-to-fine up-sampling. As can be seen from Figure \ref{network1}, 4 depth scales are outputted with $1/8,1/4,1/2$ and a full resolution. All of our loss functions are computed for each of these 4 scales, and for each of left and right images/disparities. We take the means of these loss functions as the final loss.

\textbf{Patch Size.}
By applying different patch sizes on different image scales, we can get very large equivalent patch sizes with less computation. For patch size choices, based on our empirical test, we use $n=5,5,7,9$ pixels for our patch-based ZNCC loss on 4 different scales, which is equivalent $n=5,10,28,72$ pixels' windows on full resolution images. 

\subsection{KITTI dataset.}
To be able to compare with the state-of-the-art monocular depth learning approaches, we trained and evaluated our networks using two different train/test splits: \textit{Godard} and \textit{Eigen}.

\textbf{Godard Split.}
We use the same train/test sets that Godard \textit{et al} \cite{Godard2017} proposed in their work. 200 high quality disparity images in 28 scenes provided by the official KITTI training set are served as the ground truth for benchmarking. For the rest of 33 scenes with a total of 30,159 images, 29,000 images are picked for training and the remaining 1,159 images for testing.

\textbf{Eigen Split.}
For fair comparison with more previous works, we also use the test split proposed by Eigen \textit{et al} \cite{Eigen2014} that has been widely evaluated by the works of Garg \textit{et al} \cite{Garg2016}, Liu \textit{et al} \cite{Liu2016}, Zhou \textit{et al} \cite{Zhou2017} and Godard \textit{et al} \cite{Godard2017}. This test split contains 697 images of 29 scenes. The rest of 32 scenes contain 23,488 images, in which 22,600 are used for training and the remaining for testing, similar to \cite{Garg2016} and \cite{Godard2017}.
\subsection{Results}

\subsubsection{Quantitative Evaluation.} 
The evaluation results on the KITTI dataset are reported in Table \ref{Comparison}. We use different combinations of train/test splits (E for Eigen, G for Godard) and cap distances (80m and 50m) to compare with different works. For Eigen \textit{et al} \cite{Eigen2014}, Liu \textit{et al} \cite{Liu2016}, Zhou \textit{et al} \cite{Zhou2017} and Godard \textit{et al} \cite{Godard2017} , the Eigen split with 80m cap distance are used. For Garg \textit{et al} \cite{Garg2016}, Zhou \textit{et al} \cite{Zhou2017} and Godard \textit{et al} \cite{Godard2017}, the Eigen split with 50m cap distance are used. We also report our result on Godard split with 80m cap. The results shows that our method outperforms all compared methods and produce the state-of-the-art results for monocular depth estimation problem on KITTI dataset.
\begin{table}[]
\centering
\caption{Comparison with state-of-the-art methods on KITTI dataset.}
\label{Comparison}
\scriptsize 
\setlength{\tabcolsep}{0.8pt}
\begin{tabular}{@{}lcccccccccccc@{}}
\toprule
\multirow{2}{*}{Method} & \multirow{2}{*}{\begin{tabular}[c]{@{}c@{}}Super-\\ vision\end{tabular}} &  \multirow{2}{*}{Split} & \multirow{2}{*}{Cap} & \multicolumn{5}{c}{Error (Lower better)} && \multicolumn{3}{c}{Accuracy (Higher better)}\\ \cmidrule(lr){5-9} \cmidrule(l){11-13} 
&&&& {\tiny AbsRel} & {\tiny SqRel}& {\tiny RMSE} & {\tiny RMSElog} & {\tiny D1-all}&& {\tiny$\delta <1.25$} & {\tiny$\delta <1.25^{2}$} & {\tiny$\delta <1.25^{3}$} \\ \midrule
Eigen et al \cite{Eigen2014} & Yes& E& 80 & 0.203 & 1.548 & 6.307& 0.282& - && 0.702& 0.890& 0.958\\
Liu et al \cite{Liu2016} & Yes& E& 80 & 0.201 & 1.584 & 6.471& 0.273& - && 0.680& 0.898& 0.967\\
Zhou et al \cite{Zhou2017} & No & E& 80 & 0.208 & 1.768 & 6.856& 0.283& - && 0.678& 0.885& 0.957\\
Godard et al \cite{Godard2017} & No & E& 80 & 0.148 & 1.344 & 5.927& 0.247& - && 0.803& 0.922& 0.964\\
\textbf{Ours} & \textbf{No}&  \textbf{E} & \textbf{80} & \textbf{0.145} & \textbf{1.267} & \textbf{5.786} & \textbf{0.244} & \textbf{-}&& \textbf{0.811} & \textbf{0.925} & \textbf{0.965} \\ \midrule
Garg et al \cite{Garg2016} & No & E& 50 & 0.169 & 1.080 & 5.104& 0.273& - && 0.740& 0.904& 0.962\\
Zhou et al \cite{Zhou2017} & No & E& 50 & 0.201 & 1.391 & 5.181& 0.264& - && 0.696& 0.900& 0.966\\
Godard et al \cite{Godard2017} & No & E& 50 & 0.140 & 0.976 & 4.471& 0.232& - && 0.818& 0.931& 0.969\\
\textbf{Ours} & \textbf{No}& \textbf{E} & \textbf{50} & \textbf{0.138} & \textbf{0.937} & \textbf{4.399} & \textbf{0.231} & \textbf{-}&& \textbf{0.825} & \textbf{0.933} & \textbf{0.969} \\ \midrule
Godard et al \cite{Godard2017} & No & G & 80 & 0.124 & 1.388 & 6.125& 0.217& 30.272&& 0.841& 0.936& 0.975\\
\textbf{Ours} & \textbf{No}& \textbf{G}& \textbf{80} & \textbf{0.117} & \textbf{1.202} & \textbf{5.953} & \textbf{0.210} & \textbf{29.612} && \textbf{0.845} & \textbf{0.938} & \textbf{0.976} \\ \bottomrule
\end{tabular}
\end{table}

\subsubsection{Qualitative Evaluation.} 
The qualitative comparison to some of the related methods on KITTI dataset is shown in Figure \ref{qualitativecompare}. While our network structure is similar to that of Godard \textit{et al}\cite{Godard2017}, both generate clear and accurate depth than other works. We also provide a detailed comparison with the results of Godard \textit{et al}\cite{Godard2017} in the lower part of Figure \ref{qualitativecompare}. Our network can generate more accurate depth in complex regions with thin structures and texture-less areas such as the pillars and traffic signs. This verified the theory we explained in Figure \ref{l1_zncc} that our patch-based loss function is more robust and easier to converge to the global minimum in complex regions.
{\setlength{\tabcolsep}{0.1em}
\begin{figure}
  \centering
  \scriptsize
  \begin{tabular}[c]{cccccc}
  Input & Ground-truth & Garg \textit{et al}\cite{Garg2016} & Zhou \textit{et al}\cite{Zhou2017} & Godard \textit{et al}\cite{Godard2017} & Ours \\
  \includegraphics[width=0.155\textwidth, height=0.042\textwidth]{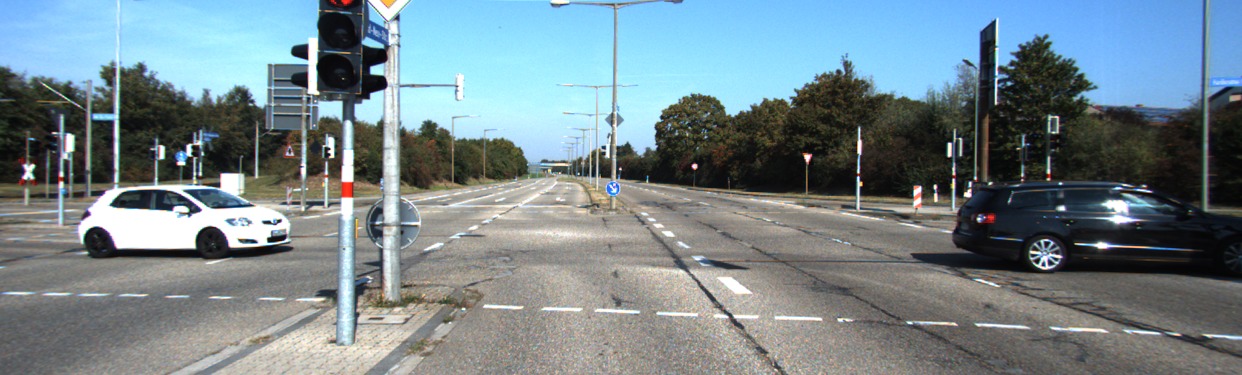} &
  \includegraphics[width=0.155\textwidth, height=0.042\textwidth]{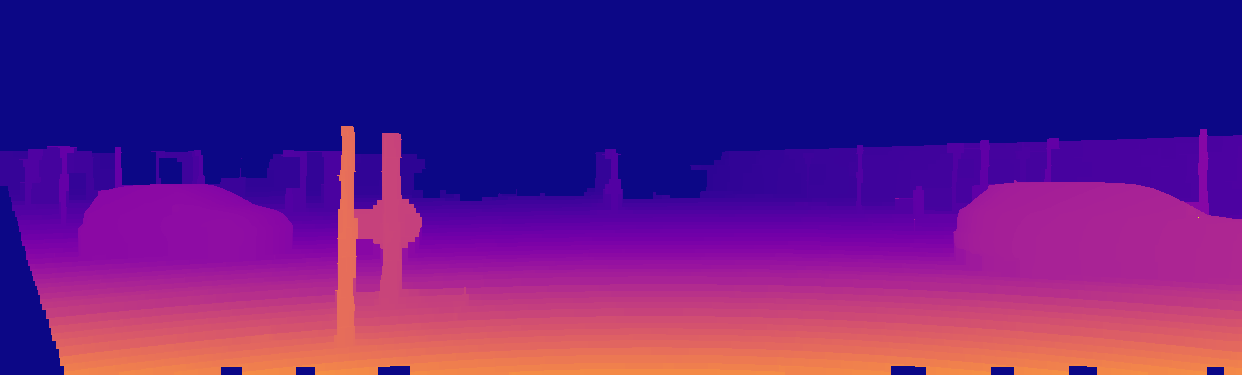} &
  \includegraphics[width=0.155\textwidth, height=0.042\textwidth]{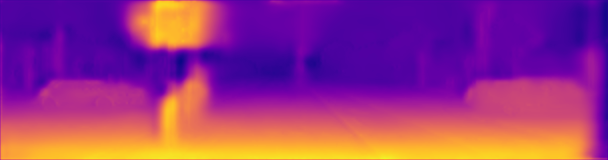} &
  \includegraphics[width=0.155\textwidth, height=0.042\textwidth]{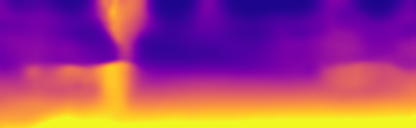} &
  \includegraphics[width=0.155\textwidth, height=0.042\textwidth]{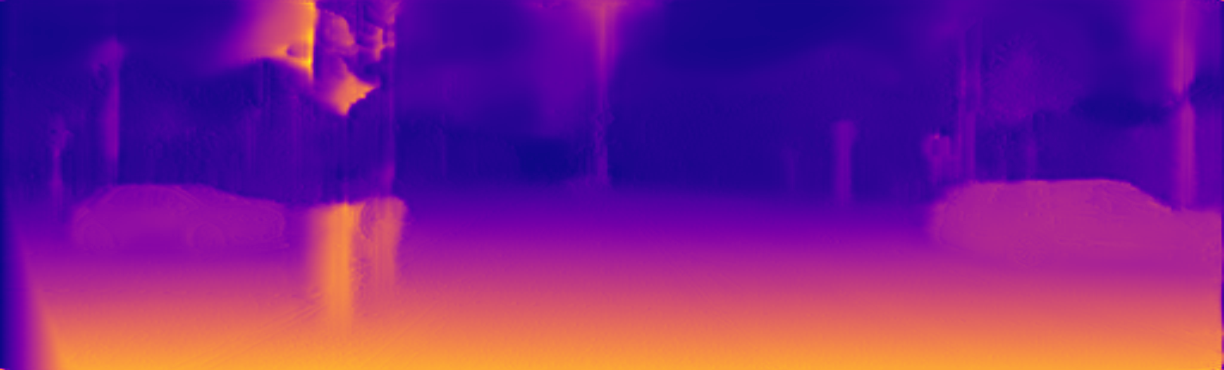} &
  \includegraphics[width=0.155\textwidth, height=0.042\textwidth]{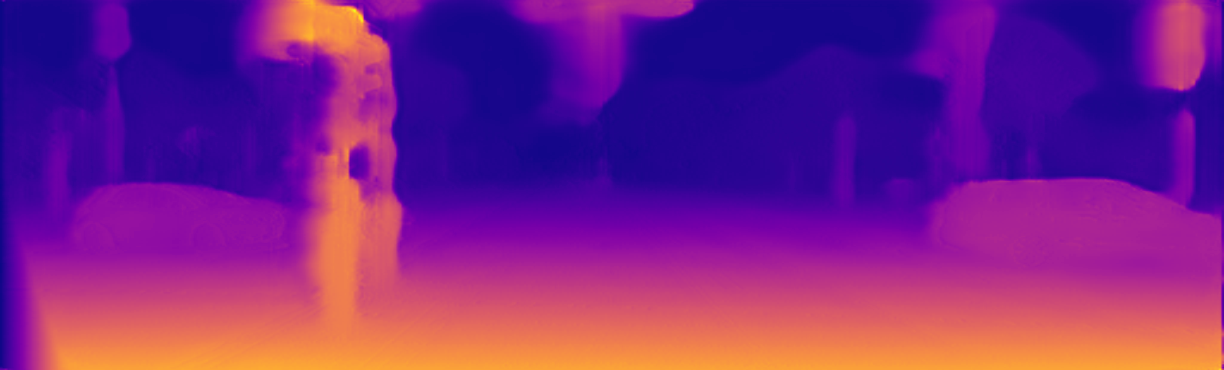}\\  
  \includegraphics[width=0.155\textwidth, height=0.042\textwidth]{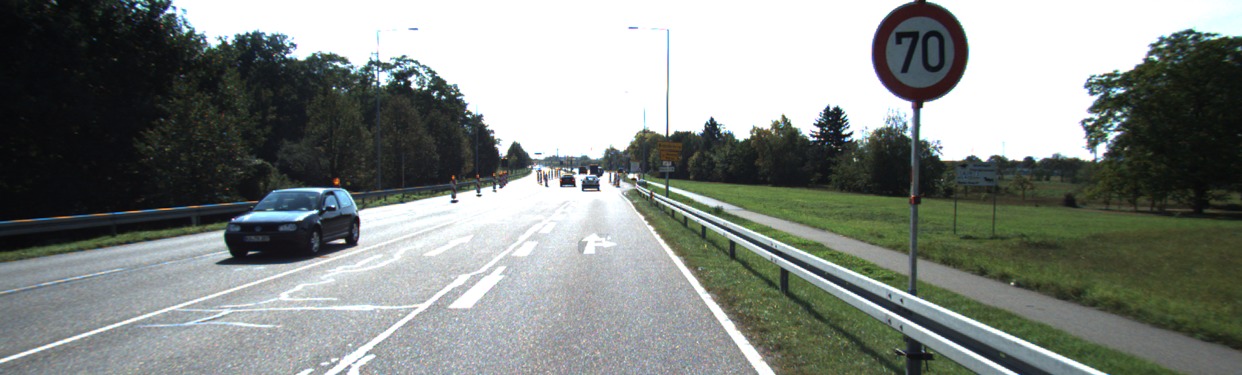} &
  \includegraphics[width=0.155\textwidth, height=0.042\textwidth]{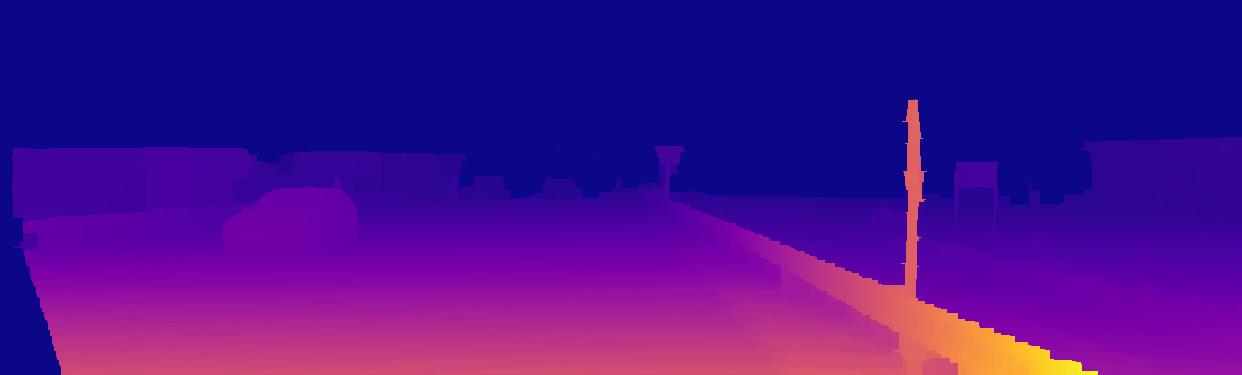} &
  \includegraphics[width=0.155\textwidth, height=0.042\textwidth]{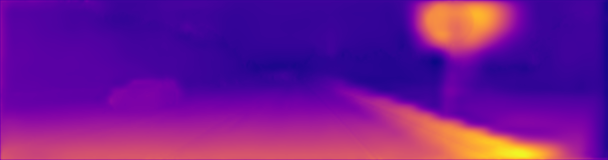} &
  \includegraphics[width=0.155\textwidth, height=0.042\textwidth]{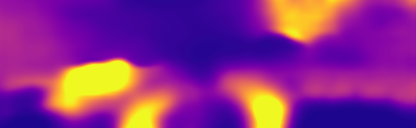} &
  \includegraphics[width=0.155\textwidth, height=0.042\textwidth]{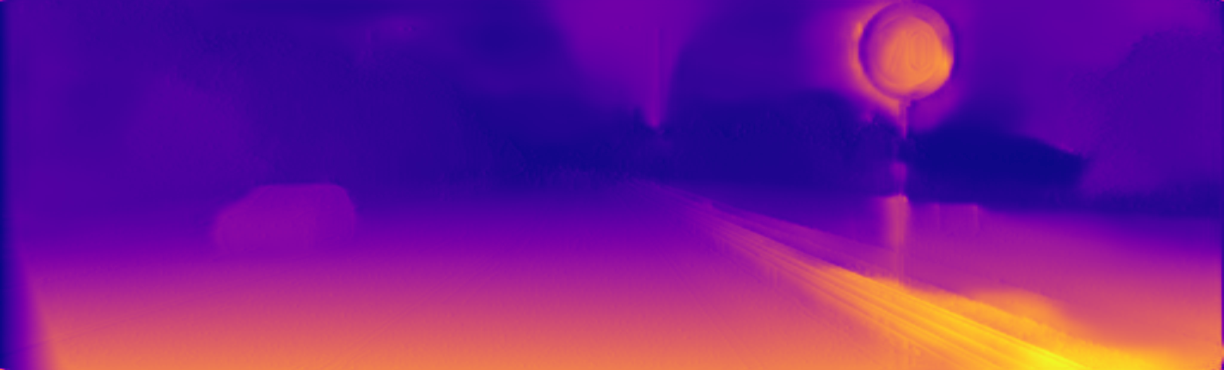} &
  \includegraphics[width=0.155\textwidth, height=0.042\textwidth]{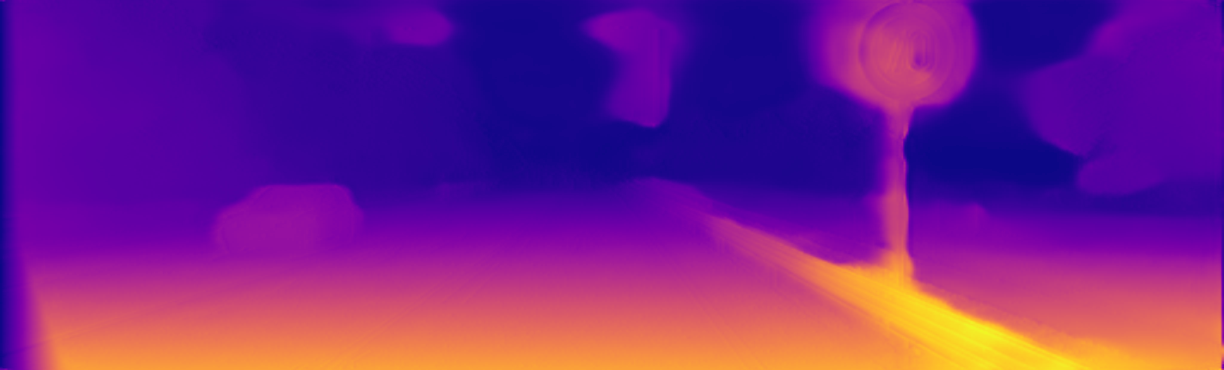}\\
  \includegraphics[width=0.155\textwidth, height=0.042\textwidth]{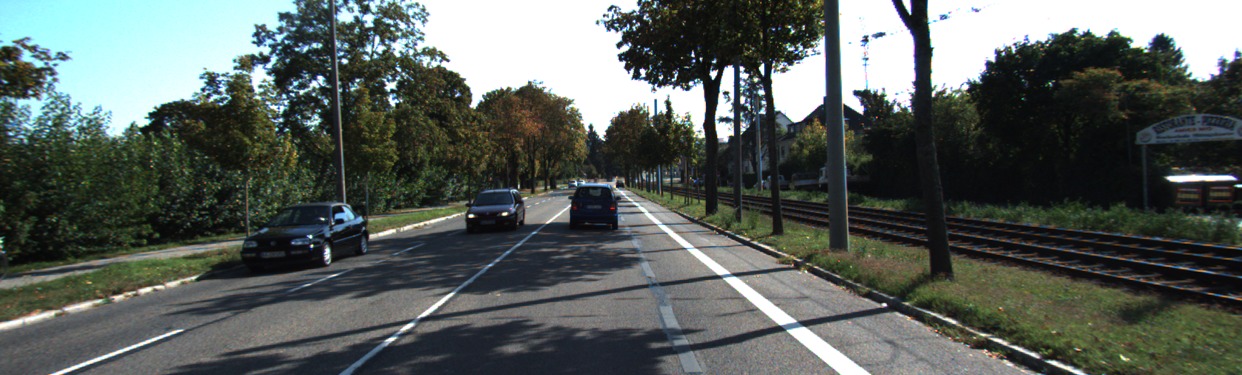} &
  \includegraphics[width=0.155\textwidth, height=0.042\textwidth]{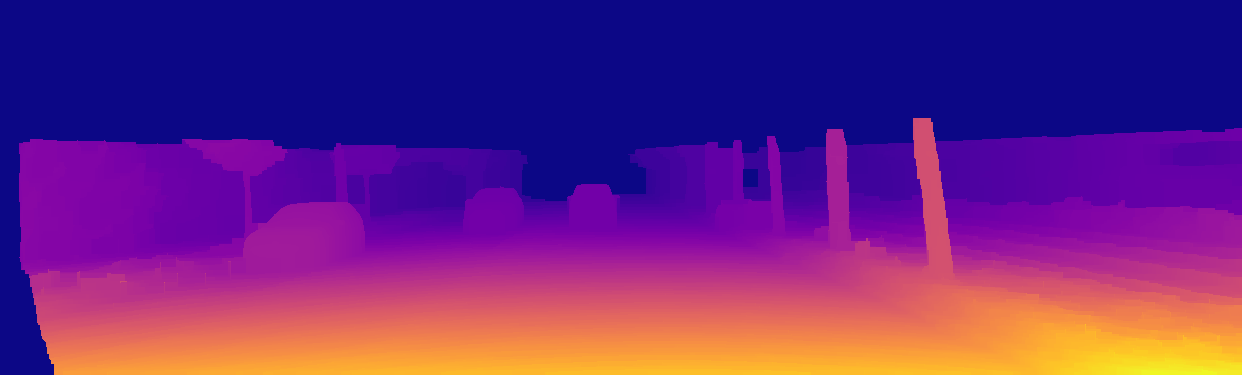} &
  \includegraphics[width=0.155\textwidth, height=0.042\textwidth]{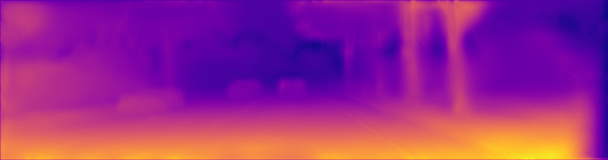} &
  \includegraphics[width=0.155\textwidth, height=0.042\textwidth]{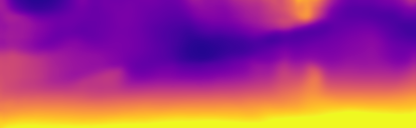} &
  \includegraphics[width=0.155\textwidth, height=0.042\textwidth]{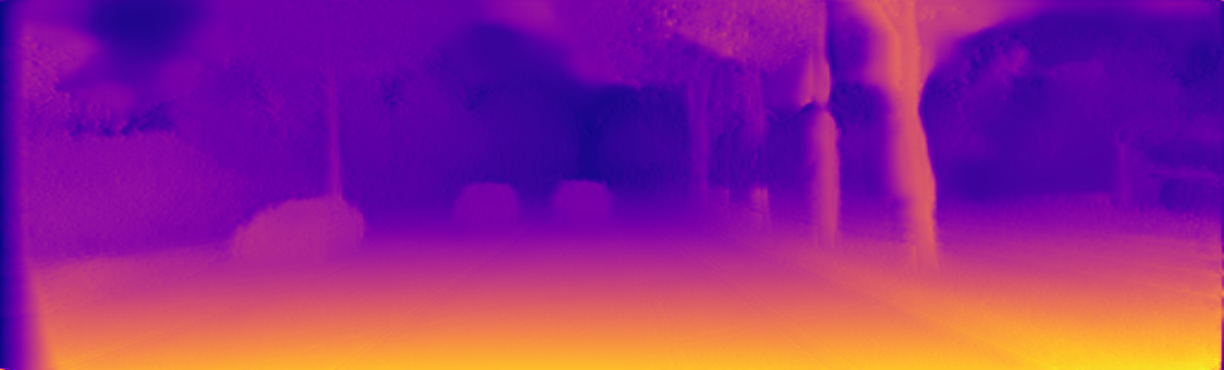} &
  \includegraphics[width=0.155\textwidth, height=0.042\textwidth]{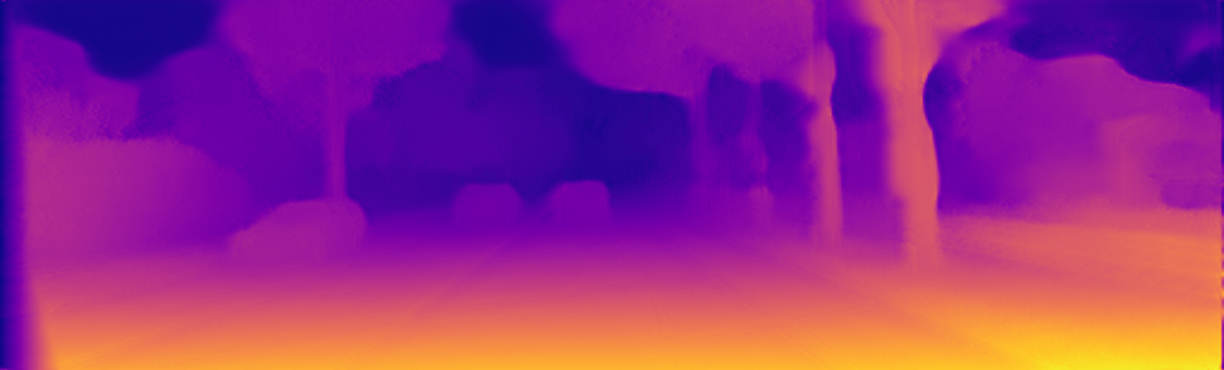} \\  
  \includegraphics[width=0.155\textwidth, height=0.042\textwidth]{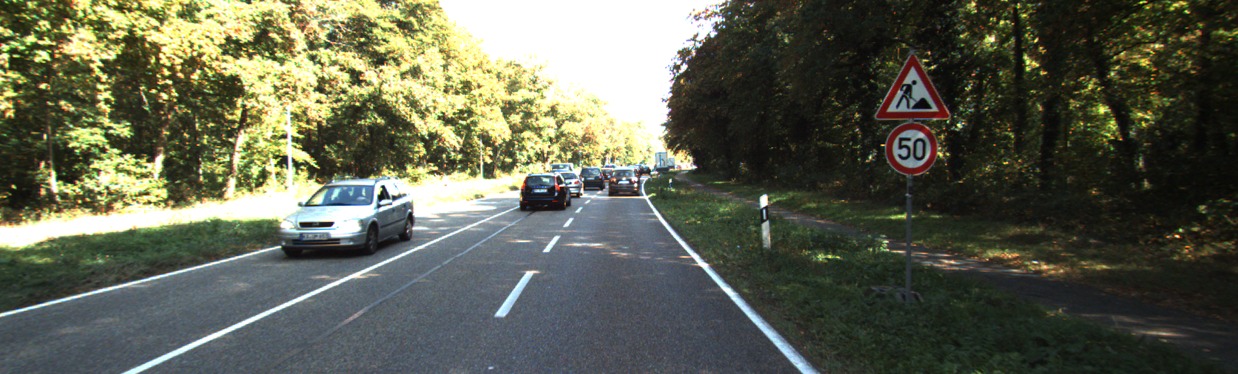} &
  \includegraphics[width=0.155\textwidth, height=0.042\textwidth]{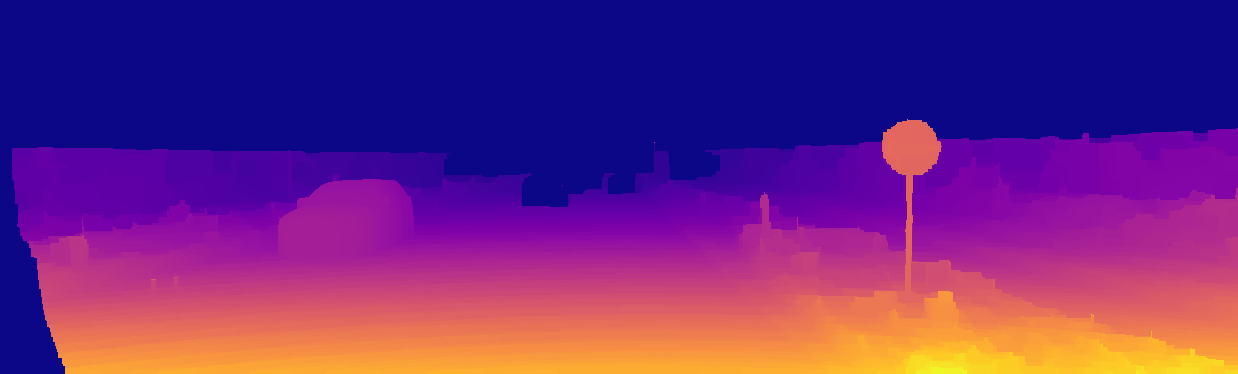} &
  \includegraphics[width=0.155\textwidth, height=0.042\textwidth]{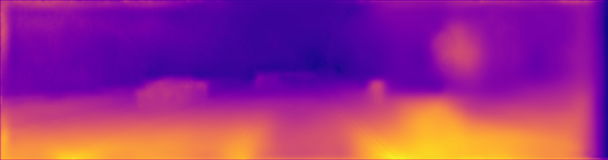} &
  \includegraphics[width=0.155\textwidth, height=0.042\textwidth]{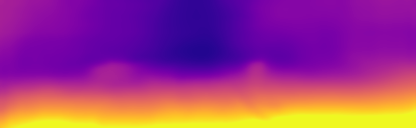} &
  \includegraphics[width=0.155\textwidth, height=0.042\textwidth]{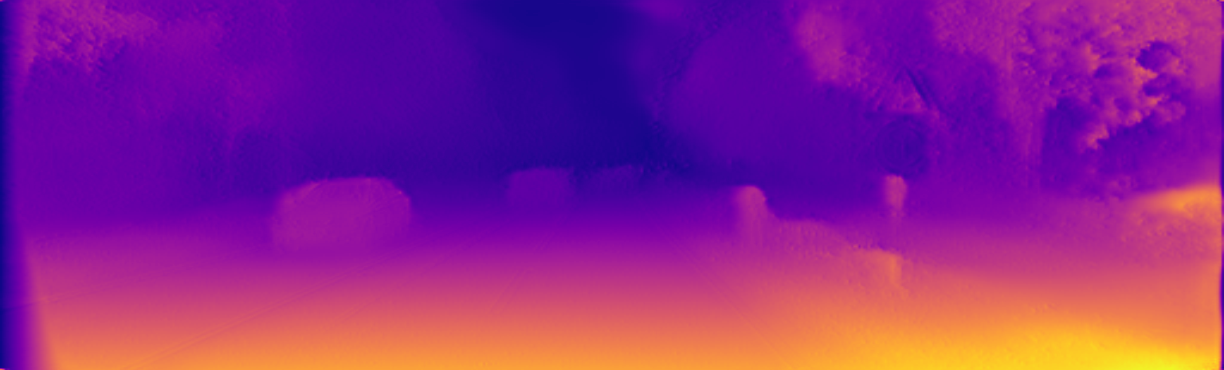} &
  \includegraphics[width=0.155\textwidth, height=0.042\textwidth]{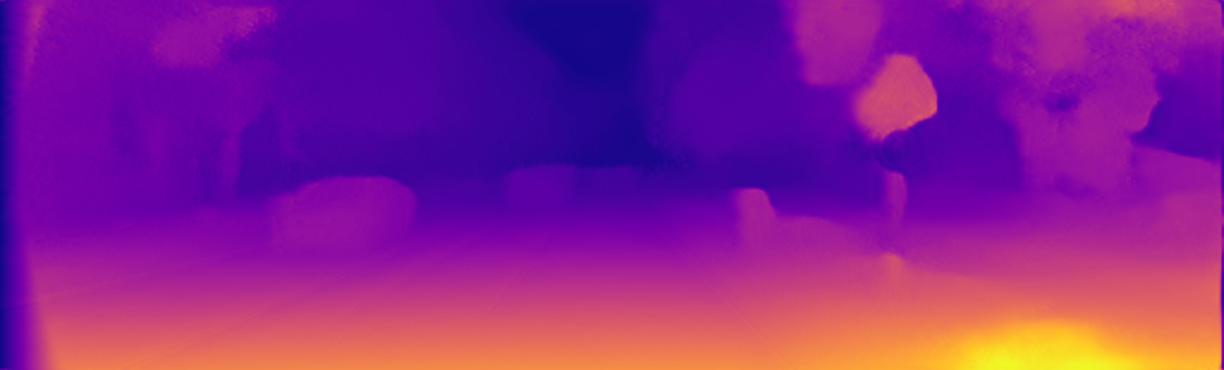}
  \end{tabular}
  \\\includegraphics[width=0.82\textwidth]{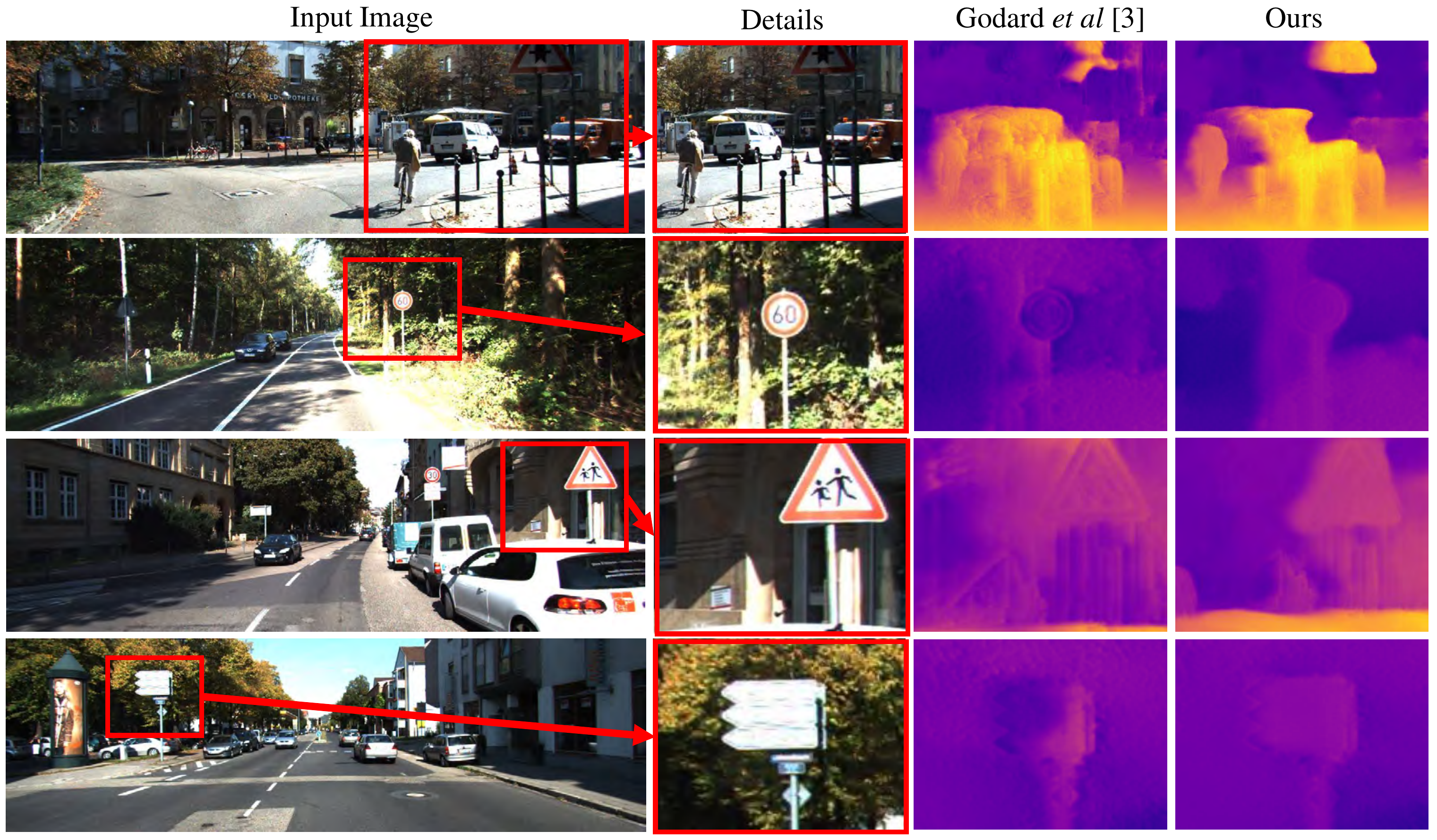}
  \caption{Upper part: comparison of monocular depth estimation on KITTI dataset between Garg \textit{et al}\cite{Garg2016}, Zhou \textit{et al}\cite{Zhou2017}, Godard \textit{et al}\cite{Godard2017}, and ours. Lower part: comparison of details with Godard \textit{et al}\cite{Godard2017}. All of the results are generated using authors' provided pre-trainned model. The ground-truth depth map is interpolated from sparse point map only for visualization.}
  \label{qualitativecompare} 
\end{figure}}

\subsubsection{Confidence Map Evaluation.} 
We show the confidence estimation results in Figure \ref{confidenceeval}. A colorbar from red to yellow is used to represent 0 to 1. We can see that the estimated confidence can nicely represent the inverted ZNCC loss but less noisy due to the small network we use to prevent over-fitting. The overlaid confidence on input image shows that our ConfidenceNet has learned to generate confidence from contextual information. For example, in texture-less areas (sky, building), dark areas (trees under shadow), occluded areas (around thin structures) and reflective areas (car window), the estimated confidence is usually very low. While the texture-rich areas and edges usually have high confidence.

{\setlength{\tabcolsep}{0.1em}
\begin{figure*}[]
  \centering
  \scriptsize
  \begin{tabular}[c]{cccccc}
  Input & Est. Depth & ZNCC Loss & Est. Confidence & Overlay \\
  \includegraphics[height=0.057\textwidth]{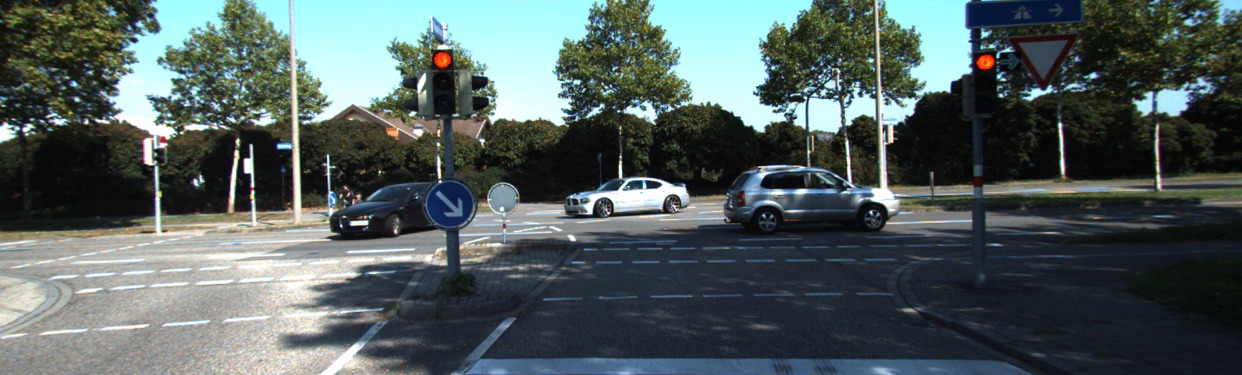} &
  \includegraphics[height=0.057\textwidth]{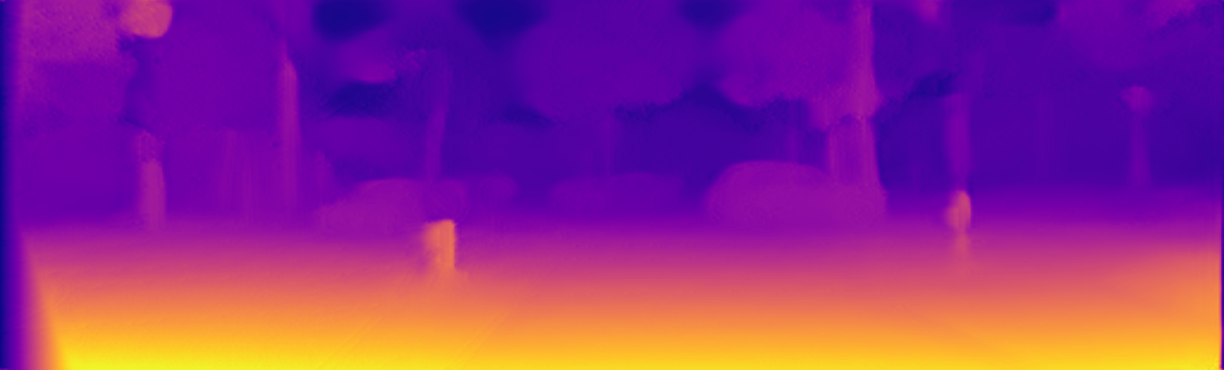} &
  \includegraphics[height=0.057\textwidth]{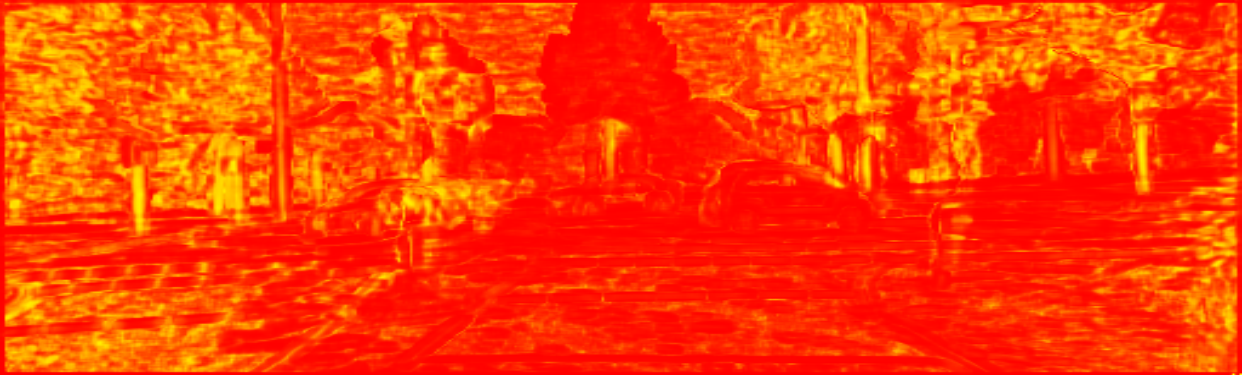} &
  \includegraphics[height=0.057\textwidth]{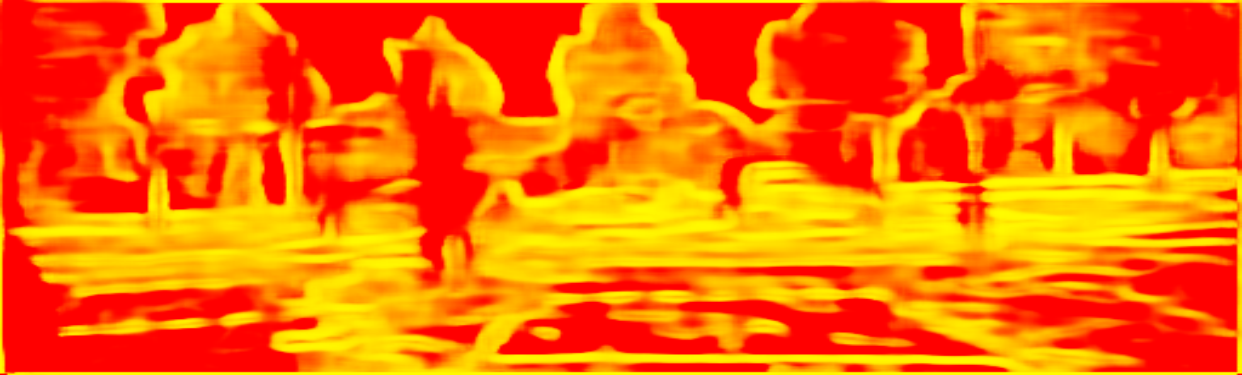} &
  \includegraphics[height=0.057\textwidth]{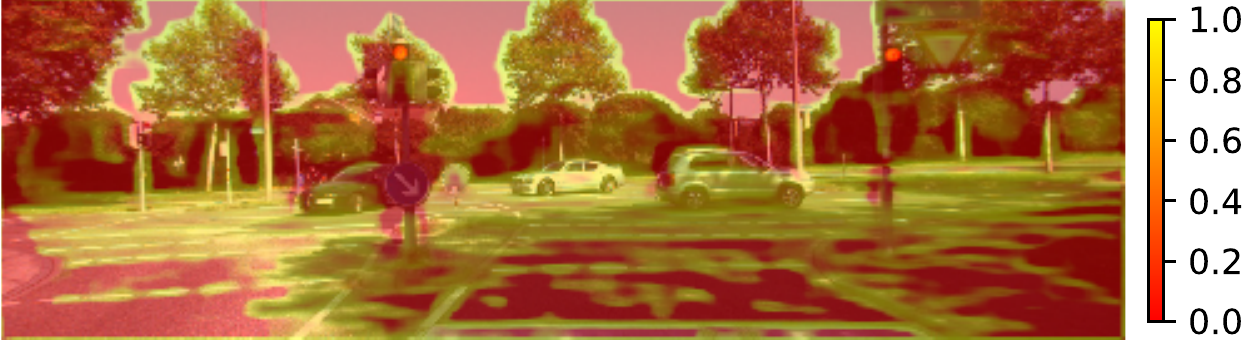}\\ 
  \includegraphics[height=0.057\textwidth]{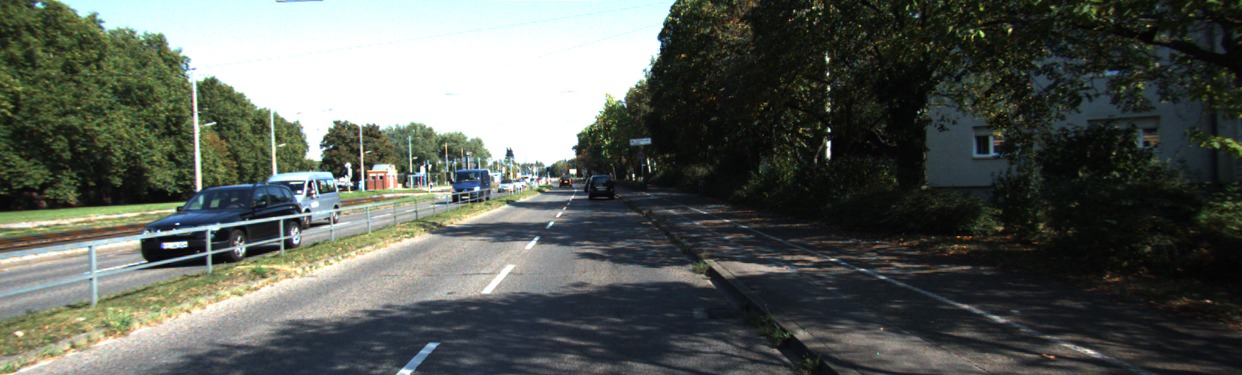} &
  \includegraphics[height=0.057\textwidth]{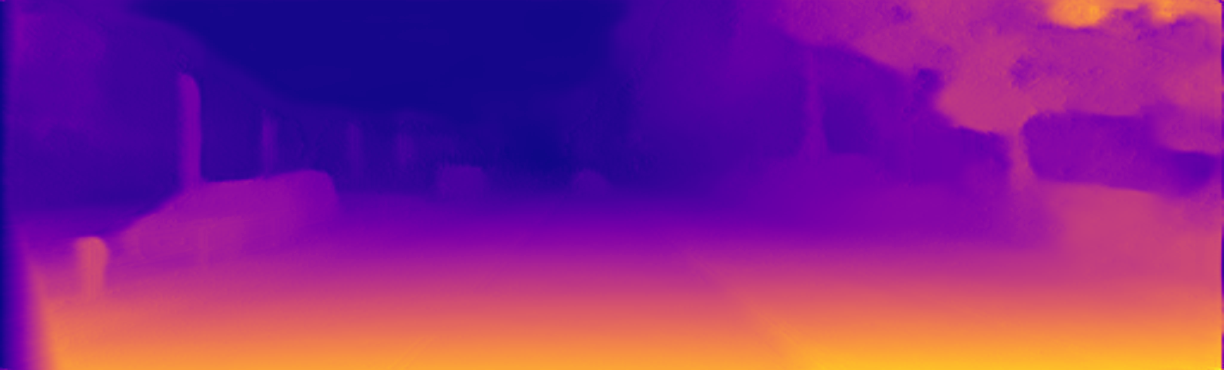} &
  \includegraphics[height=0.057\textwidth]{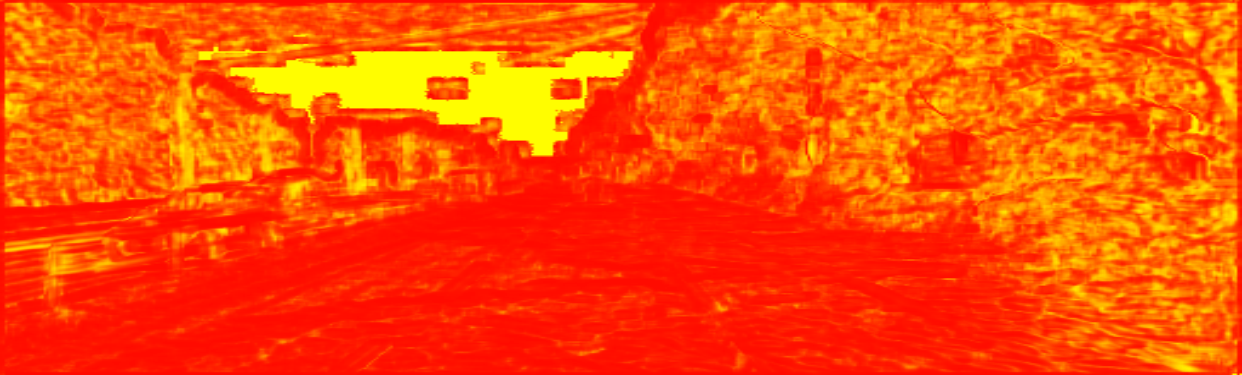} &
  \includegraphics[height=0.057\textwidth]{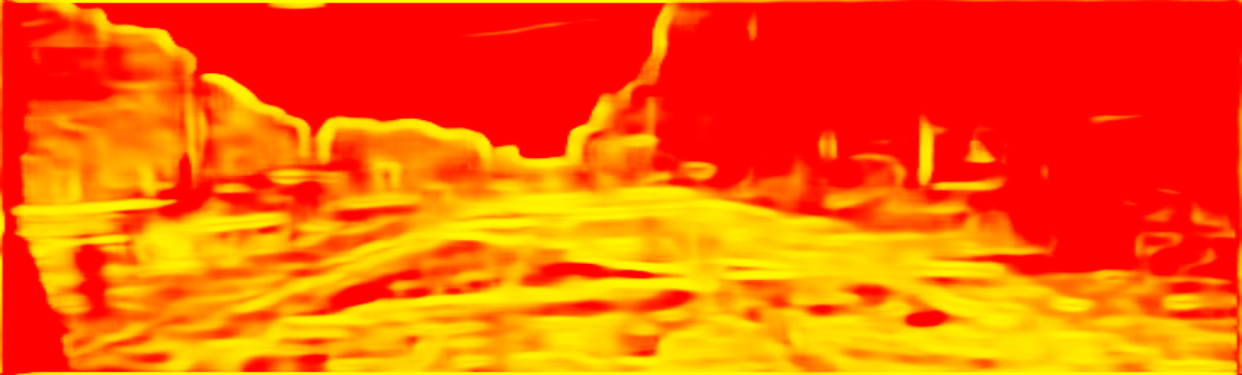} &
  \includegraphics[height=0.057\textwidth]{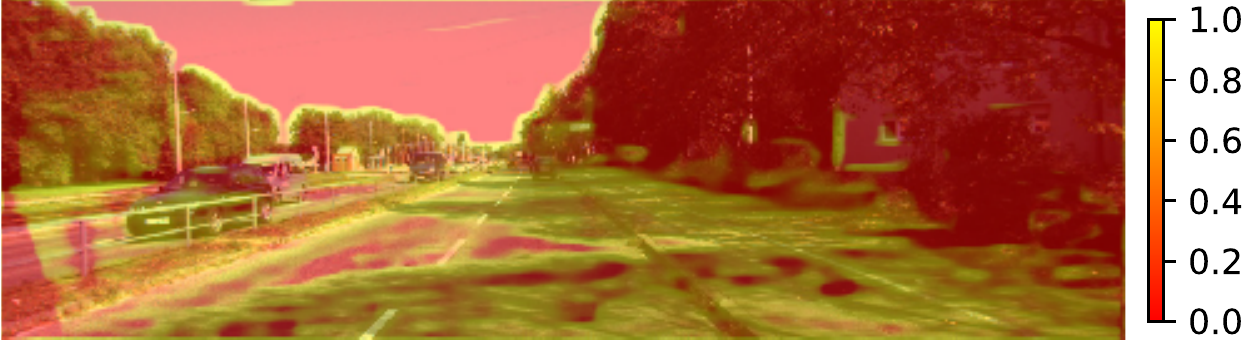}\\ 
  \includegraphics[height=0.057\textwidth]{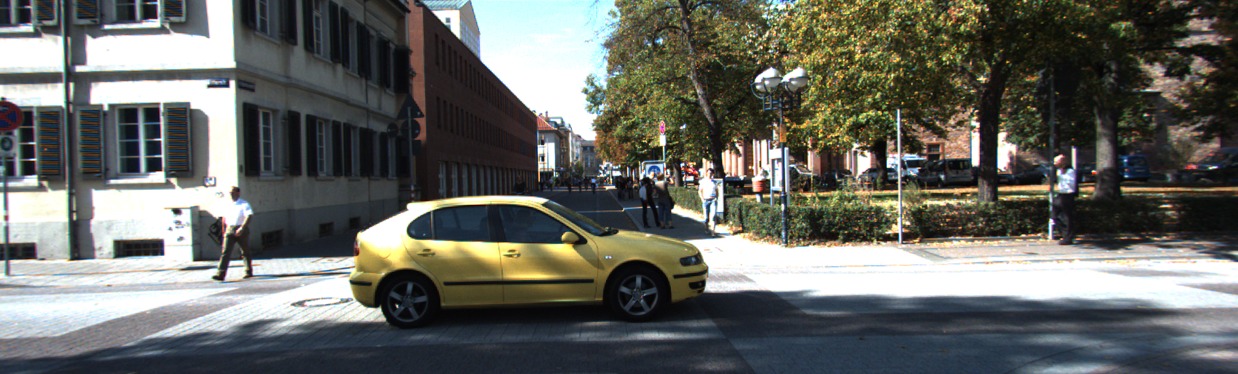} &
  \includegraphics[height=0.057\textwidth]{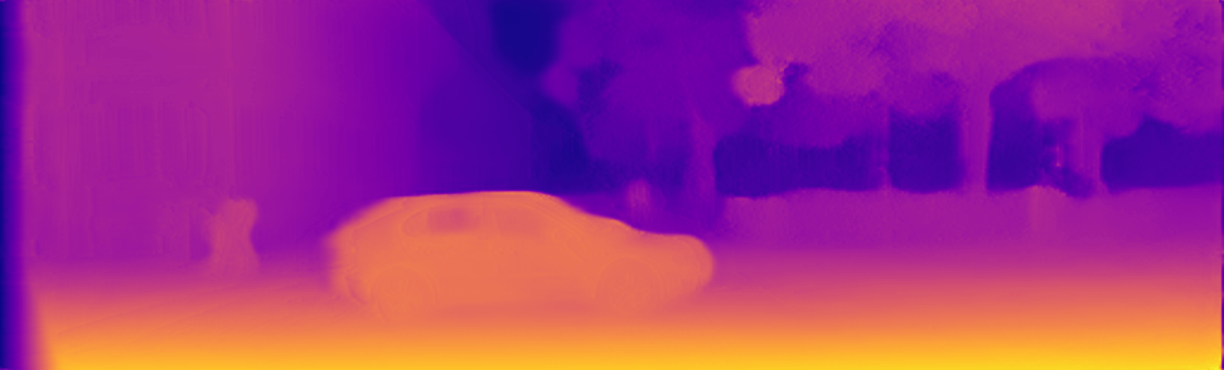} &
  \includegraphics[height=0.057\textwidth]{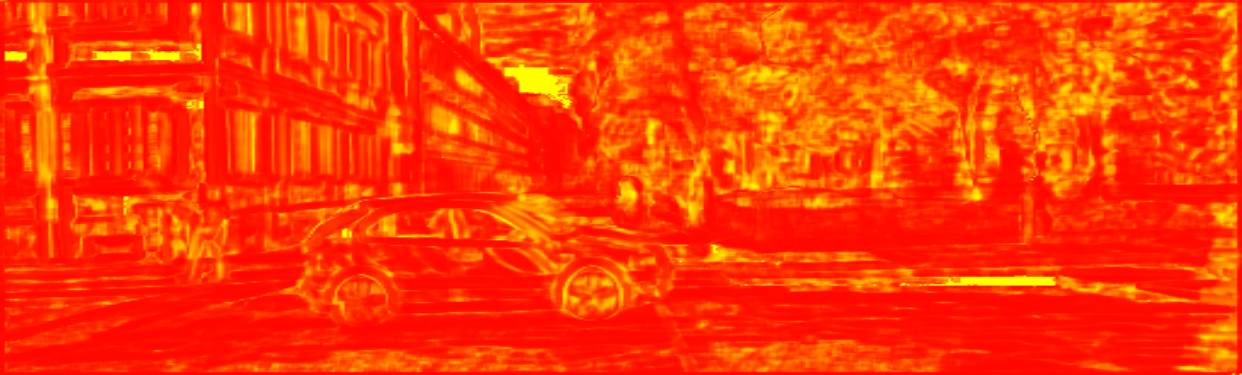} &
  \includegraphics[height=0.057\textwidth]{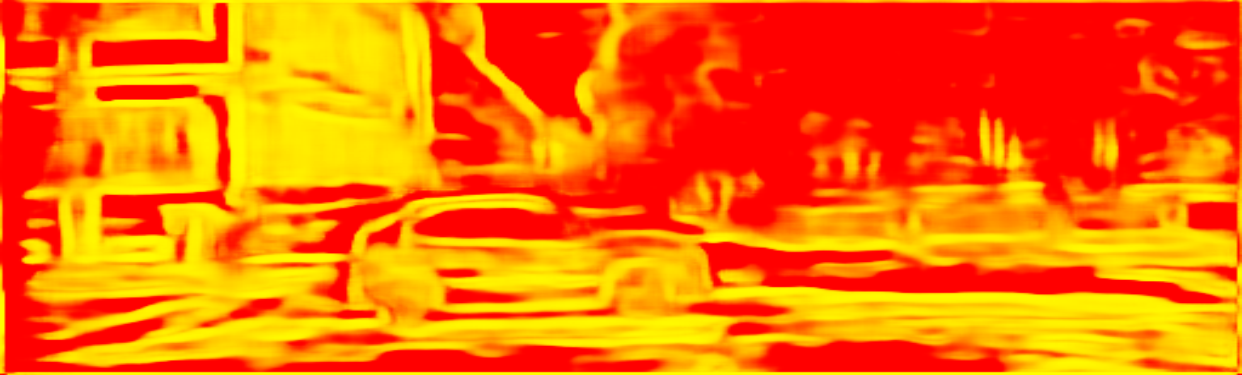} &
  \includegraphics[height=0.057\textwidth]{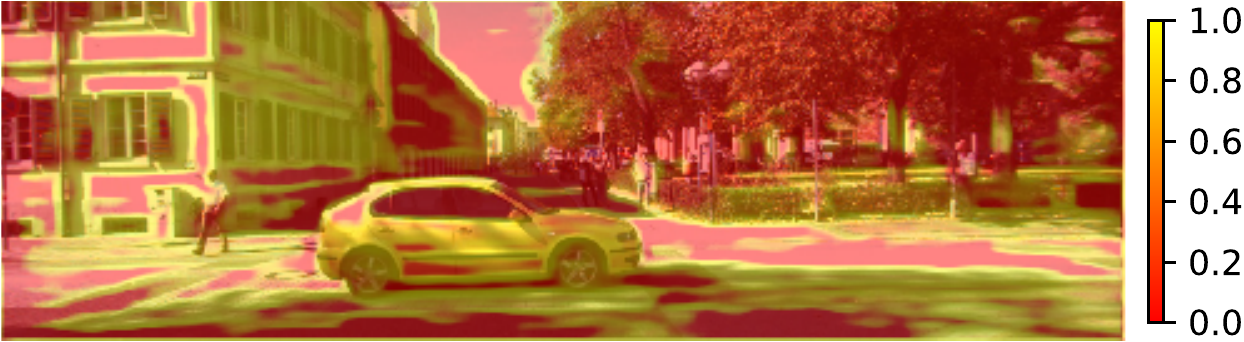}\\ 
  \includegraphics[height=0.057\textwidth]{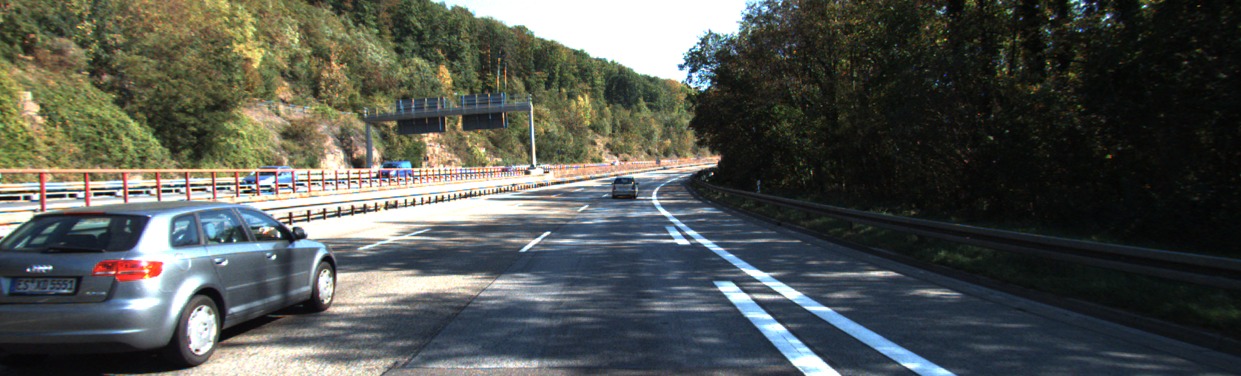} &
  \includegraphics[height=0.057\textwidth]{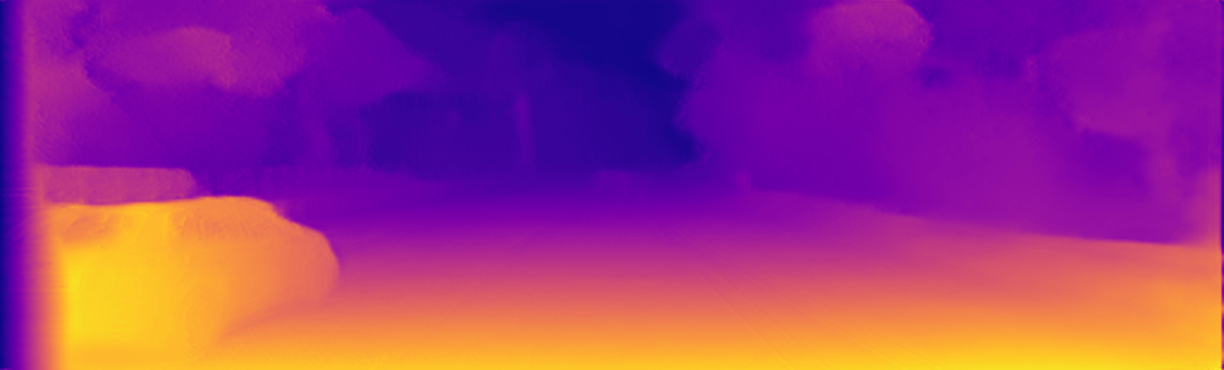} &
  \includegraphics[height=0.057\textwidth]{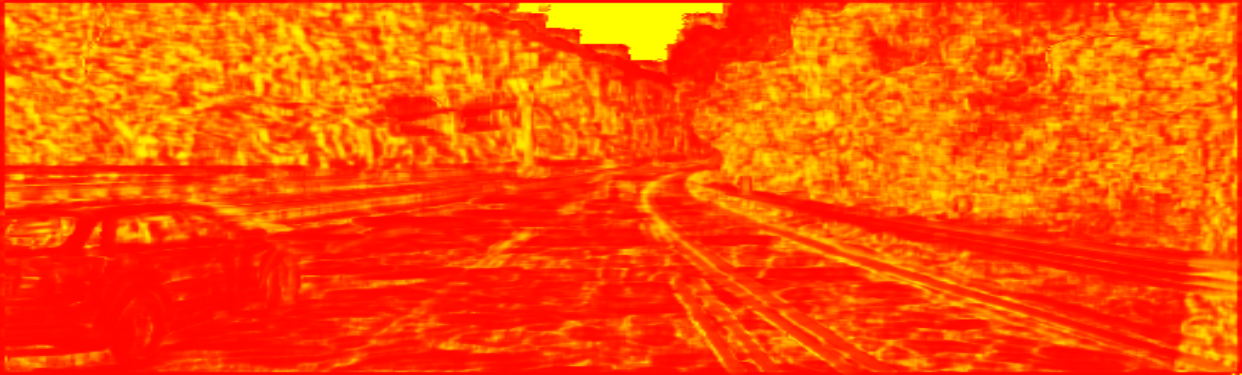} &
  \includegraphics[height=0.057\textwidth]{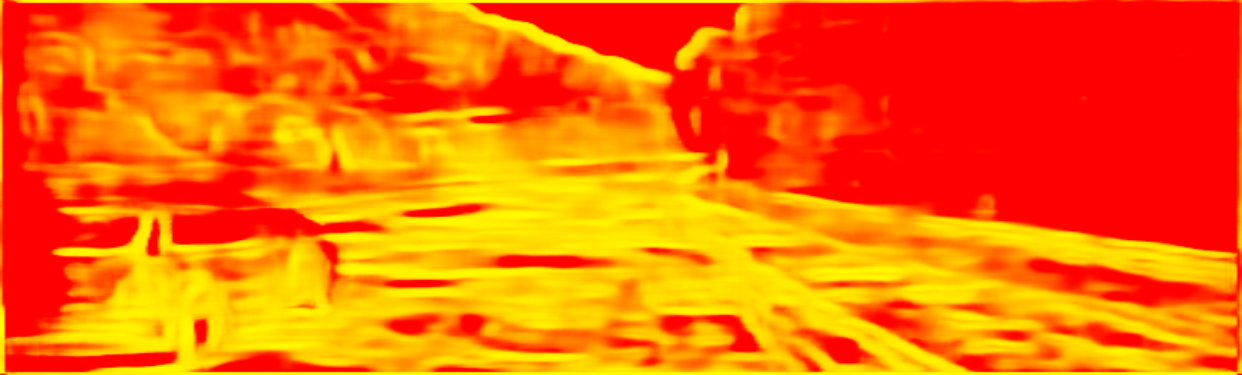} &
  \includegraphics[height=0.057\textwidth]{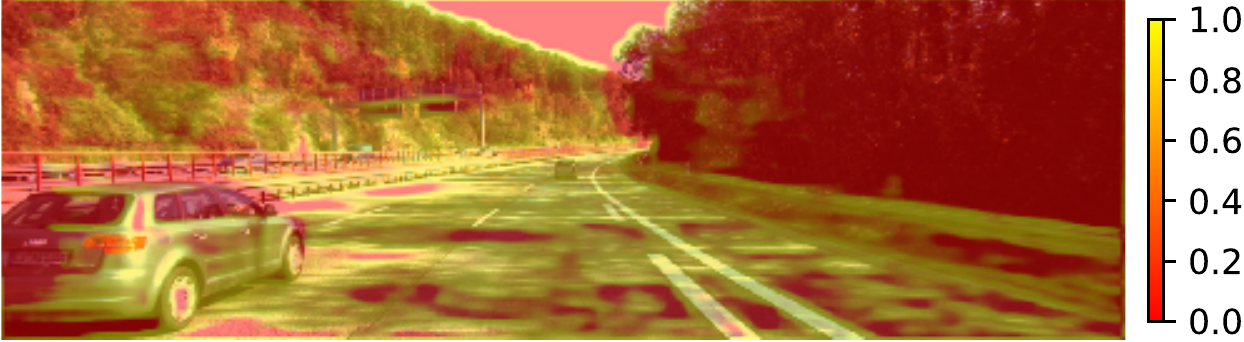}\\ 
  \end{tabular}
  \caption{Confidence estimation results. A colorbar from red to yellow is used to represent 0 to 1. }
  \label{confidenceeval} 
\end{figure*}}

\section{Discussion}
In this paper, we have presented a novel self-supervised framework for monocular depth learning and confidence estimation. We incorporate the patch matching theory into a fully differential DCNN and achieve self-supervised training of both depth and the confidence of depth. Our proposed loss function exploits the epipolar constraint of stereo vision and also provides a normalized similarity that is further used to supervise the confidence estimation. Our method not only outperforms the state-of-the-art results on the KITTI benchmark evaluation, but also for the first time, we are able to simultaneously generate depth from monocular images and estimate the confidence of the generated depth. This is a step change for monocular depth estimation as it significantly increases the feasibility of using monocular depth estimation into many practical applications such as autonomous driving and monocular endoscopic surgery, where the accuracy of estimated depth is crucial.

\textbf{Why Our ConfidenceNet Works?}
Since our ConfidenceNet is supervised by the per-pixel ZNCC loss of our depth estimation network, it explicitly learns the regions where our depth estimation network performs well and badly. But on a deeper level, our ConfidenceNet actually implicitly learns the inherent defect of the patch matching algorithm -- it would fail on texture-less regions and performs badly near stereo view occlusions, reflections and blurred areas. Therefore, after sufficient training steps, our ConfidenceNet can give an estimation of the confidence of our DepthNet, although they are two different networks.

\textbf{In Future Work.}
We will continue optimizing our model and explore the possibility of using adaptive window size for patch sampling to decrease the training time and increase accuracy in small structures.
\clearpage

\bibliographystyle{splncs}
\bibliography{egbib}

\begin{thebibliography}{10}

\bibitem{Dunkin2015}
Dunkin, B.J., Flowers, C.:
\newblock 3d in the minimally invasive surgery (mis) operating room: Cameras
  and displays in the evolution of mis.
\newblock In: Imaging and Visualization in The Modern Operating Room.
\newblock Springer (2015)  145--155

\bibitem{Krizhevsky2012}
Krizhevsky, A., Sutskever, I., Hinton, G.E.:
\newblock Imagenet classification with deep convolutional neural networks.
\newblock In Pereira, F., Burges, C.J.C., Bottou, L., Weinberger, K.Q., eds.:
  Advances in Neural Information Processing Systems 25.
\newblock Curran Associates, Inc. (2012)  1097--1105

\bibitem{Zhang1998}
Zhang, Z.:
\newblock Determining the epipolar geometry and its uncertainty: A review.
\newblock International Journal of Computer Vision \textbf{27}(2) (Mar 1998)
  161--195

\bibitem{Garg2016}
Garg, R., B.G., V.K., Carneiro, G., Reid, I.:
\newblock Unsupervised cnn for single view depth estimation: Geometry to the
  rescue.
\newblock In Leibe, B., Matas, J., Sebe, N., Welling, M., eds.: Computer Vision
  -- ECCV 2016, Cham, Springer International Publishing (2016)  740--756

\bibitem{Godard2017}
Godard, C., Aodha, O.M., Brostow, G.J.:
\newblock Unsupervised monocular depth estimation with left-right consistency.
\newblock In: 2017 IEEE Conference on Computer Vision and Pattern Recognition
  (CVPR). (July 2017)  6602--6611

\bibitem{Zhou2017}
Zhou, T., Brown, M., Snavely, N., Lowe, D.G.:
\newblock Unsupervised learning of depth and ego-motion from video.
\newblock In: 2017 IEEE Conference on Computer Vision and Pattern Recognition
  (CVPR). (July 2017)  6612--6619

\bibitem{Tateno2017}
Tateno, K., Tombari, F., Laina, I., Navab, N.:
\newblock Cnn-slam: Real-time dense monocular slam with learned depth
  prediction.
\newblock In: 2017 IEEE Conference on Computer Vision and Pattern Recognition
  (CVPR). (July 2017)  6565--6574

\bibitem{Barnard1982}
Barnard, S.T., Fischler, M.A.:
\newblock Computational stereo.
\newblock ACM Comput. Surv. \textbf{14}(4) (December 1982)  553--572

\bibitem{Scharstein2001}
Scharstein, D., Szeliski, R., Zabih, R.:
\newblock A taxonomy and evaluation of dense two-frame stereo correspondence
  algorithms.
\newblock In: Proceedings IEEE Workshop on Stereo and Multi-Baseline Vision
  (SMBV 2001). (2001)  131--140

\bibitem{Hirschmuller2008}
Hirschmuller, H.:
\newblock Stereo processing by semiglobal matching and mutual information.
\newblock IEEE Transactions on Pattern Analysis and Machine Intelligence
  \textbf{30}(2) (February 2008)  328--341

\bibitem{Kendall2017}
Kendall, A., Martirosyan, H., Dasgupta, S., Henry, P.:
\newblock End-to-end learning of geometry and context for deep stereo
  regression.
\newblock In: 2017 IEEE International Conference on Computer Vision (ICCV).
  (October 2017)  66--75

\bibitem{Eigen2014}
Eigen, D., Puhrsch, C., Fergus, R.:
\newblock Depth map prediction from a single image using a multi-scale deep
  network.
\newblock In: Proceedings of the 27th International Conference on Neural
  Information Processing Systems - Volume 2. NIPS'14, Cambridge, MA, USA, MIT
  Press (2014)  2366--2374

\bibitem{Zhang1999}
Zhang, R., Tsai, P.S., Cryer, J.E., Shah, M.:
\newblock Shape-from-shading: a survey.
\newblock IEEE Transactions on Pattern Analysis and Machine Intelligence
  \textbf{21}(8) (Aug 1999)  690--706

\bibitem{Saxena2006}
Saxena, A., Chung, S.H., Ng, A.Y.:
\newblock Learning depth from single monocular images.
\newblock In Weiss, Y., Sch\"{o}lkopf, B., Platt, J.C., eds.: Advances in
  Neural Information Processing Systems 18.
\newblock MIT Press (2006)  1161--1168

\bibitem{Saxena2007}
Saxena, A., Schulte, J., Ng, A.Y.:
\newblock Depth estimation using monocular and stereo cues.
\newblock In: Proceedings of the 20th International Joint Conference on
  Artifical Intelligence. IJCAI'07, San Francisco, CA, USA, Morgan Kaufmann
  Publishers Inc. (2007)  2197--2203

\bibitem{Saxena2008}
Saxena, A., Chung, S.H., Ng, A.Y.:
\newblock 3-d depth reconstruction from a single still image.
\newblock International Journal of Computer Vision \textbf{76}(1) (Jan 2008)
  53--69

\bibitem{Saxena2009}
Saxena, A., Sun, M., Ng, A.Y.:
\newblock Make3d: Learning {3D} scene structure from a single still image.
\newblock IEEE Transactions on Pattern Analysis and Machine Intelligence
  \textbf{31}(5) (May 2009)  824--840

\bibitem{Eigen2015}
Eigen, D., Fergus, R.:
\newblock Predicting depth, surface normals and semantic labels with a common
  multi-scale convolutional architecture.
\newblock In: 2015 IEEE International Conference on Computer Vision (ICCV).
  (December 2015)  2650--2658

\bibitem{Li2015}
Li, B., Shen, C., Dai, Y., van~den Hengel, A., He, M.:
\newblock Depth and surface normal estimation from monocular images using
  regression on deep features and hierarchical crfs.
\newblock In: 2015 IEEE Conference on Computer Vision and Pattern Recognition
  (CVPR). (June 2015)  1119--1127

\bibitem{Liu2014}
Liu, M., Salzmann, M., He, X.:
\newblock Discrete-continuous depth estimation from a single image.
\newblock In: 2014 IEEE Conference on Computer Vision and Pattern Recognition
  (CVPR). (June 2014)  716--723

\bibitem{Liu2016}
Liu, F., Shen, C., Lin, G., Reid, I.:
\newblock Learning depth from single monocular images using deep convolutional
  neural fields.
\newblock IEEE Transactions on Pattern Analysis and Machine Intelligence
  \textbf{38}(10) (October 2016)  2024--2039

\bibitem{Xu2017}
Xu, D., Ricci, E., Ouyang, W., Wang, X., Sebe, N.:
\newblock Multi-scale continuous crfs as sequential deep networks for monocular
  depth estimation.
\newblock In: 2017 IEEE Conference on Computer Vision and Pattern Recognition
  (CVPR). (July 2017)  161--169

\bibitem{Laina2016}
Laina, I., Rupprecht, C., Belagiannis, V., Tombari, F., Navab, N.:
\newblock Deeper depth prediction with fully convolutional residual networks.
\newblock In: 3D Vision (3DV), 2016 Fourth International Conference on.
  (October 2016)  239--248

\bibitem{Li2017}
Li, J., Klein, R., Yao, A.:
\newblock A two-streamed network for estimating fine-scaled depth maps from
  single rgb images.
\newblock In: 2017 IEEE International Conference on Computer Vision (ICCV).
  (October 2017)  3392--3400

\bibitem{Ladicky2014}
Ladick\'{y}, L., Shi, J., Pollefeys, M.:
\newblock Pulling things out of perspective.
\newblock In: 2014 IEEE Conference on Computer Vision and Pattern
  Recognition(CVPR). CVPR '14, Washington, DC, USA, IEEE Computer Society
  (2014)  89--96

\bibitem{Wang2015}
Wang, P., Shen, X., Lin, Z., Cohen, S., Price, B., Yuille, A.:
\newblock Towards unified depth and semantic prediction from a single image.
\newblock In: 2015 IEEE Conference on Computer Vision and Pattern Recognition
  (CVPR). (June 2015)  2800--2809

\bibitem{Mousavian2016}
Mousavian, A., Pirsiavash, H., Ko{\v{s}}eck{\'a}, J.:
\newblock Joint semantic segmentation and depth estimation with deep
  convolutional networks.
\newblock In: 3D Vision (3DV), 2016 Fourth International Conference on, IEEE
  (2016)  611--619

\bibitem{Kuznietsov2017}
Kuznietsov, Y., Stückler, J., Leibe, B.:
\newblock Semi-supervised deep learning for monocular depth map prediction.
\newblock In: 2017 IEEE Conference on Computer Vision and Pattern Recognition
  (CVPR). (July 2017)  2215--2223

\bibitem{Zoran2015}
Zoran, D., Isola, P., Krishnan, D., Freeman, W.T.:
\newblock Learning ordinal relationships for mid-level vision.
\newblock In: 2015 IEEE International Conference on Computer Vision (ICCV).
  (Dec 2015)  388--396

\bibitem{Chen2016}
Chen, W., Fu, Z., Yang, D., Deng, J.:
\newblock Single-image depth perception in the wild.
\newblock In Lee, D.D., Sugiyama, M., Luxburg, U.V., Guyon, I., Garnett, R.,
  eds.: Advances in Neural Information Processing Systems 29.
\newblock Curran Associates, Inc. (2016)  730--738

\bibitem{Cao2017}
Cao, Y., Wu, Z., Shen, C.:
\newblock Estimating depth from monocular images as classification using deep
  fully convolutional residual networks.
\newblock IEEE Transactions on Circuits and Systems for Video Technology
  \textbf{PP}(99) (2017) ~1

\bibitem{Fitzgibbon2003}
Fitzgibbon, A., Wexler, Y., Zisserman, A.:
\newblock Image-based rendering using image-based priors.
\newblock In: 2003 IEEE International Conference on Computer Vision (ICCV).
  (Oct 2003)  1176--1183 vol.2

\bibitem{Zhou2016}
Zhou, T., Tulsiani, S., Sun, W., Malik, J., Efros, A.A.:
\newblock View synthesis by appearance flow.
\newblock In: European Conference on Computer Vision. (2016)

\bibitem{Flynn2016}
Flynn, J., Neulander, I., Philbin, J., Snavely, N.:
\newblock Deep stereo: Learning to predict new views from the world's imagery.
\newblock In: 2016 IEEE Conference on Computer Vision and Pattern Recognition
  (CVPR). (June 2016)  5515--5524

\bibitem{Xie2016}
Xie, J., Girshick, R., Farhadi, A.:
\newblock Deep3d: Fully automatic 2d-to-3d video conversion with deep
  convolutional neural networks.
\newblock In Leibe, B., Matas, J., Sebe, N., Welling, M., eds.: Computer Vision
  -- ECCV 2016, Cham, Springer International Publishing (2016)  842--857

\bibitem{Luo2018}
Luo, Y., Ren, J., Lin, M., Pang, J., Sun, W., Li, H., Lin, L.:
\newblock Single view stereo matching.
\newblock In: 2018 IEEE Conference on Computer Vision and Pattern Recognition
  (CVPR). (2018)

\bibitem{Dosovitskiy2017}
Dosovitskiy, A., Springenberg, J.T., Tatarchenko, M., Brox, T.:
\newblock Learning to generate chairs, tables and cars with convolutional
  networks.
\newblock IEEE Transactions on Pattern Analysis and Machine Intelligence
  \textbf{39}(4) (April 2017)  692--705

\bibitem{Mayer2016}
Mayer, N., Ilg, E., Häusser, P., Fischer, P., Cremers, D., Dosovitskiy, A.,
  Brox, T.:
\newblock A large dataset to train convolutional networks for disparity,
  optical flow, and scene flow estimation.
\newblock In: 2016 IEEE Conference on Computer Vision and Pattern Recognition
  (CVPR). (June 2016)  4040--4048

\bibitem{Shelhamer2017}
Shelhamer, E., Long, J., Darrell, T.:
\newblock Fully convolutional networks for semantic segmentation.
\newblock IEEE Transactions on Pattern Analysis and Machine Intelligence
  \textbf{39}(4) (April 2017)  640--651

\bibitem{Jaderberg2015}
Jaderberg, M., Simonyan, K., Zisserman, A., kavukcuoglu, k.:
\newblock Spatial transformer networks.
\newblock In Cortes, C., Lawrence, N.D., Lee, D.D., Sugiyama, M., Garnett, R.,
  eds.: Advances in Neural Information Processing Systems 28.
\newblock Curran Associates, Inc. (2015)  2017--2025

\end{thebibliography}
\end{document}